\documentclass[preprint,12pt]{elsarticle}
\usepackage[utf8]{inputenc}
\usepackage{lineno,hyperref}
\usepackage{array}
\usepackage{amssymb}
\usepackage[linesnumbered,ruled,vlined]{algorithm2e}
\usepackage{enumerate}
\usepackage{makecell}
\usepackage{mathtools}
\usepackage{array}
\usepackage{multirow}
\usepackage{slashed}
\usepackage{ulem}
\usepackage{color}
\usepackage{diagbox}
\usepackage{enumerate}
\usepackage{graphicx}
\allowdisplaybreaks[1]
\usepackage{subfigure}
 \usepackage{tikz} 
\usepackage{indentfirst}
\usepackage{soul}
\usepackage{amsmath}
\usepackage[justification=centering]{caption}
\usepackage{tabularx}
\usepackage{algpseudocode}
\usepackage{adjustbox}
\usepackage{geometry}
\usetikzlibrary{shapes.geometric, arrows}
\modulolinenumbers[5]

\usepackage{epstopdf}
\usepackage{booktabs}
\usepackage{tabularx}
\usepackage{array}
\usepackage{lscape}
\usepackage{ulem}
\usepackage{amssymb,amsmath}
\usepackage{longtable}
\usepackage{booktabs}
\usepackage{setspace}
\usepackage{threeparttable}
\usepackage{epsfig}
\usepackage{colortbl}
\usepackage{amsmath,amsthm,amssymb,amsfonts}
\usepackage[utf8]{inputenc}
\usepackage{epsfig,graphics,subfigure,psfrag,amsmath,amssymb}

\biboptions{comma,square}
\usepackage{amsmath}
\usepackage{threeparttable}
\usepackage{multirow}
\usepackage{mathtools}
\usepackage{amsmath}
\usepackage{tikz}

\makeatletter
\thm@headfont{\sc}
\makeatother
\usepackage{adjustbox}

\usepackage{float}
\usepackage{amsthm}

\usepackage{enumerate}

\usepackage{enumerate}
\usepackage{properties}
\usepackage{amsmath, amssymb}
\usepackage{makecell}
\usepackage{multirow}
\usepackage{rotating}

\newcommand{\circledsmall}[1]{\lower.7ex\hbox{\tikz\draw (0pt, 0pt)%
    circle (.5em) node {\makebox[0.1em][c]{\small#1}};}}

\newcommand{\circledtiny}[1]{\lower.7ex\hbox{\tikz\draw (0pt, 0pt)%
    circle (.3em) node {\makebox[0.1em][c]{\tiny #1}};}}

\journal{Information Sciences}

\begin{document}

\begin{frontmatter}

\title{An Order-Sensitive Conflict Measure for Random Permutation Sets}

\author[inst1,inst2]{Ruolan Cheng}
\author[inst1,inst3]{Yong Deng\corref{label1}}
\cortext[label1]{Corresponding author.\\ Email address: dengentropy@uestc.edu.cn(Yong Deng).
}

\affiliation[inst1]{organization={Institute of Fundamental and Frontier Science},
organization={University of Electronic Science and Technology of China},
            city={Chengdu},
            postcode={610054},
            country={China}}
\affiliation[inst2]{organization={Andalusian Research Institute in Data Science and Computational Intelligence},
organization={Dept. of Computer Science and AI, University of Granada},
            city={Granada},
            postcode={18071},
            country={Spain}}
\affiliation[inst3]{organization={School of Medicine},
organization={Vanderbilt University},
            city={Nashville},
            postcode={37240},
            country={USA}}

\begin{abstract}
Random permutation set (RPS) is a new formalism for reasoning with uncertainty involving order information. Measuring the conflict between two pieces of evidence represented by permutation mass functions remains an open issue in order-dependent uncertain information fusion. This paper analyzes conflicts in RPS from two different perspectives: random finite set (RFS) and Dempster-Shafer theory (DST). From the DST perspective, the order information incorporated into focal sets reflects a qualitative propensity where higher-ranked elements are more significant. Motivated by this view and observations on permutations, we define a non-overlap-based inconsistency measure for permutations and develop an order-sensitive conflict measure for RPSs. The proposed method reformulates the conflict in RPSs as a graded, order-dependent notion rather than a simple dichotomy of conflict versus non-conflict. Numerical examples are presented to validate the behavior and properties of the proposed conflict measure. The proposed method not only exhibits an inherent top-weightedness property and effectively quantifies conflict between RPSs within the DST framework, but also provides decision-makers with flexibility in selecting weights, parameters, and truncation depths.

\end{abstract}


\end{frontmatter}

\section{Introduction}
Uncertainty is an inherent characteristic of information. It arises from a lack of comprehensive knowledge regarding a specific problem, leaving agents, whether human decision-makers or machines, unable to accurately describe the environment or predict its future evolution \cite{cuzzolin2020geometry}. Effective knowledge representation and reasoning under such uncertain environments are fundamental to the advancement of explainable artificial intelligence. Consequently, the modeling and reasoning mechanisms for uncertain information in intelligent systems have consistently been a focal point of research. To address this, numerous mathematical theories have emerged, including interval probability \cite{keynes2013treatise}, subjective probability \cite{anscombe1963definition}, fuzzy set theory \cite{zadeh1965fuzzy}, Dempster-Shafer theory \cite{dempster2008upper}, fuzzy measure theory \cite{sugeno1974theory}, possibility theory \cite{dubois2012possibility}, and Z-numbers \cite{zadeh2011note}. These frameworks are widely applied across various fields, such as risk assessment, machine learning, artificial intelligence, engineering design, and strategic planning \cite{cheng2022novel,xiao2024complex,li2025z}.

Dempster–Shafer theory (DST) of evidence, as a non-additive probability measure, gained popularity for its ability to model ignorance arising from insufficient information or data presented in the form of sets. It generalizes the basic event space of classical probability theory to the power set and traditional probability distribution to basic probability assignment (BPA). To measure the uncertainty of the power set information distribution, Deng\cite{deng2016deng} proposed a novel belief entropy, known as Deng entropy, and explained the power set from the perspective of entropy. Under this interpretation, power sets are regarded as potential combinations of all basic events\cite{song2021entropic}, which naturally prompts an intriguing question: \textit{What mathematical structure is formed by the permutations of these basic events?} Inspired by this inquiry, Deng\cite{deng2022random} introduced order information into propositions, proposing a new set concept called random permutation set (RPS). RPS theory (RPST) further extends the power set space in DST to the permutation event space (PES), which is the set of permutation states of all basic events, and constructs the permutation mass function (PMF). It expresses higher-dimensional uncertain information, breaking through the unorderedness of the set and providing a theoretical framework for processing uncertain information with ordered information. 

Although RPST remains in its infancy, its theoretical framework has evolved rapidly. In the seminal RPS literature \cite{deng2022random}, Deng introduced the left intersection (LI) and right intersection (RI) operations for ordered focal sets, and extended Dempster's combination rule to the left orthogonal sum (LOS) and right orthogonal sum (ROS). Subsequent studies have further enriched RPST in several aspects, including uncertainty measures~\cite{zhou2025limit,li2025improved,liu2026fractal}, distance and divergence measures~\cite{chen2023distance,chen2023permutation,chen2024symmetric}, and information fusion mechanisms ~\cite{zhou2023conjunctive,zhou2024order,wang2024new,su2026bdos,deng2024random,li2025combining}. Meanwhile, its theoretical foundations have been further strengthened through matrix-based implementations of core operations~\cite{yang2023matrix}, marginalization methods~\cite{zhou2023marginalization}, and investigations into its information fractal dimension~\cite{zhao2023information,liu2025multifractal}. Despite these advances, a key issue remains unresolved: conflict measure in RPST. When intelligent systems process information from multiple sources, inconsistent evidence often occurs due to incompletely reliable sources or divergent interpretations, leading to conflicts. Understanding, characterizing, and quantifying this conflict is vital when combining information from multiple sources. The concept and quantification of conflicts when information is represented by belief functions have been extensively discussed\cite{destercke2013toward,pichon2019several,liu2006analyzing,abellan2021combination,bronevich2022measures,burger2016geometric}. In DST, conflict is usually used as an indicator to select appropriate combination rules, discount certain information that lacks reliability, or support decision-making. However, in RPST, conflict has received only a cursory mention in the original RPS literature\cite{deng2022random}. Deng defines left conflict $\overset{\leftarrow}{K}$ and right conflict $\overset{\rightarrow}{K}$ based on the LI and RI operations, where the conflict between ordered focal sets is determined by whether LI or RI is empty. Unlike traditional sets, such judgment conditions cannot effectively reflect order conflicts between permutations. For example, the LI or RI of $(\omega_3\omega_2\omega_1)$ and $(\omega_1\omega_2)$ are both non-empty, but it cannot be concluded that there is no conflict between them because the order of the elements is obviously different. This also leads to information loss caused by the inability of existing RPS fusion rules\cite{deng2022random,zhou2023conjunctive,wang2024new,su2026bdos} to maximize the utilization of the order information in the ordered focal set. The conflict measure between RPSs remains a critical research gap that needs to be urgently addressed.

Beyond the above limitation of intersection-based conflict measures, the semantic interpretation of order information in propositions also deserves careful consideration. Order is a pervasive attribute in many real-world settings, including temporal sequences, causal relations, action execution, and magnitude comparison. It specifies relative relationships among elements, and even slight mismatches in order may lead to substantially different conclusions. Hence, accurately representing and managing order information is essential for reliable reasoning and decision-making. In this respect, Zhou et al.~\cite{zhou2023conjunctive} provided an important semantic foundation by clarifying information distribution in RPS from both RFS and DST perspectives. Specifically, they introduced a layer-2 belief-structure interpretation, where order is viewed as a qualitative propensity that reflects the decision-maker’s tendency of belief transfer. Under this interpretation, a key property of an ordered focal set is that higher-ranked elements carry greater importance. This insight has further supported subsequent studies on RPS reasoning~\cite{deng2024random}, clustering~\cite{chen2025occm}, and generation~\cite {chen2026ddhrps}. It also suggests that conflict measures should be developed under a specific interpretation of information distribution, rather than defined solely by empty/non-empty intersections.

To this end, this paper conducts an in-depth analysis of the conflict between RPSs from two distinct perspectives, RFS and DST, with a primary focus on the issue of RPS conflict measure under the DST view. Starting from observations of permutations, we first define a non-overlap-based conflict measure between permutations, then extend it to permutation mass functions, and finally develop an order-sensitive conflict measure for RPSs. The proposed method reformulates conflict in RPSs as an order-sensitive measure, rather than a binary judgment based solely on the presence of an intersection between focal elements. Numerical examples are presented to illustrate the behavior and properties of the proposed method. The results show that the proposed measure effectively exploits order information in ordered focal sets and overcomes limitations of existing approaches. In addition, it naturally exhibits a top-weighted characteristic and allows decision-makers to flexibly choose weights, parameters, and truncation depths according to practical requirements and preferences. Therefore, it provides a useful tool for handling conflict in future multi-source information fusion tasks involving order-uncertain information.

The rest of this paper is organized as follows. Section \ref{2} reviews the basic concepts involved in DST, RPST, and subsequent conflict measures. An order-sensitive conflict measure method on RPS is proposed in Section \ref{3}. Section \ref{4} discusses the behavior and properties of the proposed conflict measure in detail with numerical examples. Finally, a summary is provided in Section \ref{5}.

\section{Preliminaries}
\label{2}
RPS, as a new concept proposed in 2022, is still little known, so we start from the DST to introduce the RPS. In addition, some basic concepts related to conflict measures in the following are also given in this section.

\subsection{Dempster-Shafer theory}

$\mathbf{Definition \ 2.1.}$ \textit{(Frame of discernment\cite{shafer1976mathematical})} Frame of discernment(FoD) $\Omega$ is an exhaustive set of all possible values of the variable $\mathcal{W}$. If the number of elements of $\Omega$ is $n$, it can be represented as

\begin{equation}
  \label{eq1}
\Omega=\{\omega_1, \omega_2, ..., \omega_n\}.
\end{equation}
And the elements in $\Omega$ are mutually exclusive.

The power set on $\Omega$ is denoted as $\mathcal{P}(\Omega)$, which contains all possible combinations of propositions in $\Omega$, a total of $2^n$.
\begin{equation}\label{eq2}
\begin{aligned}
\mathcal{P}(\Omega)&=\{F_0, F_1, F_2, ..., F_i, ..., F_{2^n-1}\}\\
&=\{\emptyset, \{\omega_1\},\{\omega_2\}, ...,\{\omega_n\}, \{\omega_1, \omega_2\}, \{\omega_1, \omega_3\}, ..., \{\omega_1, \omega_n\}, ...,\Omega \}.
\end{aligned}
\end{equation}

 \begin{table}[!htbp]
    	\begin{center}
        		\caption{Encoding of focal sets in DST and ordered focal sets in RPST}
                \begin{tabular}{ccccccc}
                 \hline
                 \toprule
                 &  &\multicolumn{3}{c}{Encoding}&  &
                    \\ \cmidrule(lr){3-5}
    			$\mathcal{F}$ & $\mathcal{OF}$ & Binary &Decimal &Order& $F_i$  & $F_i^j$   \\
               \hline
                 \specialrule{0em}{0.5pt}{0.5pt}
    			$\emptyset$ & $\emptyset$ & 0...000 &0 & 1	&$F_0$ &$F_0^1$ \\
        		\specialrule{0em}{0.5pt}{0.5pt}
    			$\{\omega_1\}$         &    $(\omega_1)$     & 0...001 &1 & 1 &$F_1$ &$F_1^1$  \\
        		\specialrule{0em}{0.5pt}{0.5pt}
    			$\{\omega_2\}$          &     $(\omega_2)$     & 0...010 &2 & 1 &$F_2$ &$F_2^1$  \\
        		\specialrule{0em}{2pt}{2pt}
                 \multirow{2}{*}{$\{\omega_1\omega_2\}$}    & $(\omega_1\omega_2)$     &  \multirow{2}{*}{0...011}   &\multirow{2}{*}{3}   &1&\multirow{2}{*}{$F_3$} &$F_3^1$ \\
                                            & $(\omega_2\omega_1)$     &                             &                     &2&                       &$F_3^2$ \\
        		\specialrule{0em}{2pt}{2pt}
    			$\{\omega_3\}$          &      $(\omega_3)$     & 0...100 &4 & 1 &$F_4$ &$F_4^1$  \\
                \specialrule{0em}{2pt}{2pt}
                 \multirow{2}{*}{$\{\omega_1\omega_3\}$}    & $(\omega_1\omega_3)$     &  \multirow{2}{*}{0...101}   &\multirow{2}{*}{5}   &1&\multirow{2}{*}{$F_5$} &$F_5^1$ \\
                                            & $(\omega_3\omega_1)$     &                             &                     &2&                       &$F_5^2$ \\
                \specialrule{0em}{2pt}{2pt}
                 \multirow{2}{*}{$\{\omega_2\omega_3\}$}    & $(\omega_2\omega_3)$     &  \multirow{2}{*}{0...110}   &\multirow{2}{*}{6}   &1&\multirow{2}{*}{$F_6$}        &$F_6^1$ \\
                                            & $(\omega_3\omega_2)$     &                             &                     &2&                              &$F_6^2$ \\
                \specialrule{0em}{2pt}{2pt}
                 \multirow{4}{*}{$\{\omega_1\omega_2\omega_3\}$}    & $(\omega_1\omega_2\omega_3)$     &  \multirow{4}{*}{0...111}   &\multirow{4}{*}{7}    &1&\multirow{4}{*}{$F_7$} &$F_7^1$ \\
                                                                    & $(\omega_1\omega_3\omega_2)$     &                             &                      &2&                       &$F_7^2$ \\
                                                                    & $\vdots$                         &                             &                      &$\vdots$    &                       &$\vdots$ \\
                                                                    & $(\omega_3\omega_2\omega_1)$     &                             &                      &6&                       &$F_7^6$ \\
                 \specialrule{0em}{2pt}{2pt}
    			$\vdots$  &$\vdots$  &$\vdots$  &$\vdots$  & $\vdots$ 	&$\vdots$  &$\vdots$  \\
                \specialrule{0em}{2pt}{2pt}
                 \multirow{4}{*}{$\{\Omega\}$}                      & $(\omega_1\omega_2...\omega_n)$      &  \multirow{4}{*}{1...111}   &\multirow{4}{*}{$2^n-1$}    &1&\multirow{4}{*}{$F_{2^n-1}$} &$F_{2^n-1}^1$ \\
                                                                    & $(\omega_1...\omega_n\omega_{n-1})$                            &                             &                            & 2  & &$F_{2^n-1}^2$ \\
                                                                    & $\vdots$                             &                             &                            & $\vdots$   &                  &$\vdots$ \\
                                                                    & $(\omega_n\omega_{n-1}...\omega_1)$  &                             &                            &$n!$    &    &$F_{2^n-1}^{n!}$ \\
    			\bottomrule
    			\label{tab1}
    		\end{tabular}
    	\end{center}
    \end{table}

$\mathbf{Definition \ 2.2.}$ \textit{(Mass function\cite{shafer1976mathematical})} A mass function $m$ on $\Omega$, also known as the Basic Belief Assignment(BBA), is a mapping from the power set $\mathcal{P}(\Omega)$ to the unit interval $[0, 1]$, $m: \mathcal{P}(\Omega) \rightarrow [0, 1]$, with $\sum_{F_i\in \mathcal{P}(\Omega)}m(F_i)=1$. In Shafer's original literature\cite{shafer1976mathematical}, the condition $m(\emptyset)=0$ must be satisfied. But, there are also other perspectives, especially in the TBM\cite{smets1994transferable}, where the condition $m(\emptyset)=0$ is often omitted. $m(\emptyset)>0$ usually indicates conflicts or unknown targets in the open world. If there is a subset $F_i$ of $\Omega$ such that $m(F_i)>0$, then $F_i$ is called a focal set of $m$. Each focal set of $\Omega$ can be encoded by a unique binary\cite{zhou2023generating}, as shown in Table \ref{tab1}, where 1 and 0 indicate the presence and absence of the element, respectively. $i$ in $F_i$ is the decimal conversion of the binary encoding of the focal set. The focal set can be uniquely determined by $i$, for example, $F_5 \Leftrightarrow \{\omega_1, \omega_3\}$. $\mathcal{F}$ denotes the set of all focal sets, and the body of evidence is written as $<\Omega, \mathcal{F}, m>$. 

Suppose $m_1$ and $m_2$ are two mass functions defined on $\Omega$ respectively from two reliable and independent sources.

$\mathbf{Definition \ 2.3.}$ \textit{(Conflict of mass functions\cite{shafer1976mathematical})} The conflict $k$ between $m_1$ and $m_2$ is numerically equal to the sum of the products of the beliefs assigned to the focal set whose intersection is empty.
\begin{equation}
    \label{eq3}
k=\sum_{F_i\cap F_j=\emptyset}m_1(F_i)m_2(F_j)
\end{equation}
It plays a vital role in normalization operations in Dempster’s combination rule.

\subsection{Random permutation set theory}

RPS\cite{deng2022random} is a new formalism for reasoning with uncertainty involving order information. It is a higher-dimensional uncertainty information processing framework based on PES, and the information processing unit is PMF. Some basic concepts in RPST are as follows.

$\mathbf{Definition \ 2.4.}$ \textit{(Permutation event space\cite{deng2022random})} The permutation event space in RPS contains all possible permutations of propositions in $\Omega$, denoted as $\mathcal{PES}(\Omega)$.
\begin{equation}\label{eq6}
\begin{aligned}
\mathcal{PES}(\Omega)&=\{F_0^1, F_1^1, F_2^1, ..., F_i^j, ..., F_{2^n-1}^{n!}\}\\
&=\{\emptyset, (\omega_1),(\omega_2), ...,(\omega_n), (\omega_1, \omega_2), (\omega_2, \omega_1), ..., (\omega_{n-1}, \omega_n),  \\
&\ \ \ \ \  (\omega_{n}, \omega_{n-1}), ..., (\omega_1, \omega_2, ..., \omega_n), ..., (\omega_n, \omega_{n-1}, ..., \omega_1) \}
\end{aligned}
\end{equation}
When $|\Omega|=n$, $|\mathcal{PES}(\Omega)|=\sum_{k=0}^{n}P(n, k)=\sum_{k=0}^{n}\frac{n!}{(n-k)!}$.

$\mathbf{Definition \ 2.5.}$ \textit{(Permutation mass function\cite{deng2022random})} A permutation mass function $Perm$ on $\Omega$ is a mapping from permutation event space $\mathcal{PES}(\Omega)$ to the unit interval $[0, 1]$, $Perm: \mathcal{PES}(\Omega) \rightarrow [0, 1]$ with $\sum_{F_i^j\in \mathcal{PES}(\Omega)}Perm(F_i^j)=1$. When $Perm(F_i^j)>0$, $F_i^j$ is the ordered focal set of $Perm$, and $\mathcal{OF}$ represents the set of all ordered focal sets. Similarly, the body of evidence in RPS is written as $<\Omega, \mathcal{OF}, Perm>$. In Table \ref{tab1}, the 2-tuple $(i, j)$ is used to encode the ordered focal set, where $i$ plays the same role as it does in DST, and $j$ indicates the order of $F_i$. Each ordered focal set in $\mathcal{PES}(\Omega)$ can be uniquely determined by $(i, j)$\cite{zhou2023conjunctive}, for example, $F_5^2 \Leftrightarrow \{\omega_3, \omega_1\}$.

Suppose $Perm_1$ and $Perm_2$ are two permutation mass functions defined on $\Omega$ respectively from two reliable and independent sources.

$\mathbf{Definition \ 2.6.}$ \textit{(Left and right conflict of permutation mass functions\cite{deng2022random})} The left conflict $\overset{\leftarrow}{K}$ and the right conflict $\overset{\rightarrow}{K}$ between $Perm_1$ and $Perm_2$ are respectively formally defined as
\begin{equation}
    \label{eq7}
\overset{\leftarrow}{K}= \sum_{F_i^m\overset{\leftarrow}{\cap}F_j^n=\emptyset}Perm_1(F_i^m)Perm_2(F_j^n),
\end{equation}

\begin{equation}
    \label{eq8}
\overset{\rightarrow}{K}= \sum_{F_i^m\overset{\rightarrow}{\cap}F_j^n=\emptyset}Perm_1(F_i^m)Perm_2(F_j^n).
\end{equation}
Where $\overset{\leftarrow}{\cap}$ and $\overset{\rightarrow}{\cap}$ are the left and right intersection of permutation events, respectively, defined as

\begin{equation}
    \label{eq9}
F_i^m\overset{\leftarrow}{\cap}F_j^n=F_i^m\backslash \underset{\omega\in F_i^m,\omega\not\in F_{j}^{n}}{\bigcup}\{\omega\},
\end{equation}

\begin{equation}
    \label{eq10}
F_i^m\overset{\rightarrow}{\cap}F_j^n= F_{j}^{n}\backslash \underset{\omega\in F_{j}^{n},\omega\not\in F_i^m}{\bigcup}\{\omega\}.
\end{equation}
The symbol $"\backslash"$ denotes an asymmetric set-difference operation, which means removing the latter from the former. The left and right operations imply that the result obeys the order information of the side pointed by the arrow.

\subsection{Rank-biased overlap}
Rank-biased overlap (RBO) is an effective technique for measuring the similarity of indefinite ordered lists characterized by non-conjointness, top-weightedness, and incompleteness. RBO was first proposed by Webber et al.\cite{webber2010similarity} in 2010 and has subsequently been used in various fields such as public search engines\cite{rieder2018ranking,urman2022matter,makhortykh2020search}, retrieval systems\cite{abdelrazek2023topic,moffat2017incorporating,fernandez2022multistage}, and machine learning\cite{liu2020deep,bai2016sparse}. It measures the similarity of two lists by calculating the intersection size of sets corresponding to different depths of the lists.

Let $S$ and $T$ be two ordered lists. $S_i$ is the $i$th element of $S$, $S_{c: d}=\{S_i: c\le i\le d\}$ is the set of elements from the $c$-th position to the $d$-th position of the ordered list. When the depth is $d$, the intersection of $S$ and $T$ is denoted as
\begin{equation}
    \label{eq13}
I_d=S_{1:d}\cap T_{1:d}.
\end{equation}
The cardinality of the intersection $I_d$ is called the overlap of the ordered lists $S$ and $T$ at depth $d$, denoted $|I_d|$. The ratio of overlap $|I_d|$ to depth $d$ reflects the degree of consistency between lists $S$ and $T$.
\begin{equation}
    \label{eq14}
C_d=\frac{|I_d|}{d}=\frac{|S_{1:d}\cap T_{1:d}|}{d}
\end{equation}

$\mathbf{Definition \ 2.7.}$ \textit{(Average overlap\cite{wu2003methods})} For each $d\in \{1,2,...,n\}$, calculate the overlap at $d$. As the name implies, average overlap is the average of these overlaps:
\begin{equation}
    \label{eq15}
AO(S, T, n)=\frac{1}{n}\sum_{d=1}^{n}C_d,
\end{equation}
where $n$ is the depth to be calculated.

$\mathbf{Definition \ 2.8.}$ \textit{(Rank biased overlap\cite{webber2010similarity})} Different from the average overlap, RBO is no longer simply average cumulative overlap but assigns a weight $\alpha$ to the consistency of each different depth and then calculates the weighted sum.
\begin{equation}
    \label{eq16}
SIM(S, T, \alpha)=\sum_{d=1}^{\infty}\alpha_d\cdot C_d
\end{equation}

Let the weight be $\alpha_d=(1-p)\cdot p^{d-1}$, $p\in(0, 1)$. Since when $0<p<1$, the geometric series $p^{d-1}$ converges to $\frac{1}{1-p}$, that is
\begin{equation}
    \label{eq17}
\sum_{d=1}^{\infty}p^{d-1}=\frac{1}{1-p}.
\end{equation}
Hence, there are $\sum_{d}\alpha_d=1$. According to the definition $0\leq C_d\leq1$, then there is $0\le SIM\le \sum_{d}\alpha_d\leq1$.

At this point, the RBO similarity measure method can be written as
\begin{equation}
    \label{eq18}
RBO(S, T, p)=(1-p)\sum_{d=1}^{\infty}p^{d-1}\cdot C_d.
\end{equation}
It is easy to prove that the value of RBO falls in the range $[0, 1]$, where 0 represents two completely different lists, and 1 represents consistent lists. The parameter $p$ indicates the decrease in weight. The smaller $p$ is, the greater the top weight is, and the steeper the weight declines. When $p$ takes a limit value of 0, it means that only the frontmost element of the list is considered. On the contrary, if $p$ takes a value of 1, the weights are arbitrarily flat but meaningless.

\section{An order-sensitive conflict measure for Random Permutation Sets}
\label{3}
In this section, we discuss the conflict in RPS from RFS and DST perspectives and point out the problems of the existing RPS conflict measure method. Furthermore, an order-sensitive conflict measure method from the DST view is defined, and the calculation process of the proposed method is demonstrated through a numerical example.

\subsection{Discussion on the conflict in RPS from different perspectives}
Probability theory models the randomness of events, while DST extends the event space in probability theory to power sets, modeling imprecision arising from limited knowledge by assigning beliefs to multi-element focal sets. In addition, the Random Finite Set, which is also a power set distribution, considers that the multi-element focal set is generated by the simultaneous observation of two random variables, the type and cardinality of the element. Recently, Zhou et al.\cite{zhou2023conjunctive} provided different explanations of PMF in RPST from both DST and RFS perspectives(\cite{zhou2023conjunctive}, Fig. 1). From the perspective of RFS, the order of elements is the newly added third random variable. An ordered focal set is generated by three random variables: type, cardinality, and order of the element. In the threat assessment case presented in \cite{deng2022random}, a radar directly detects the permutation of enemy aircraft; here, distinct permutations constitute unique events, aligning with the RFS paradigm. To clarify these distinctions, we present a concrete example below.

$\mathbf{Example \ 3.1.}$ Let $RPS_1$, $RPS_2$ and $RPS_3$ from three distinct sources on the frame $\Omega=\{\omega_1, \omega_2, \omega_3\}$ , and their PFMs are respectively
\begin{equation}
\begin{aligned}
&PMF_1(\Omega): Perm_1(F_3^1)=1\ \ \Rightarrow\ Perm_1((\omega_1\omega_2))=1, \\
&PMF_2(\Omega): Perm_2(F_7^1)=1\ \ \Rightarrow\ Perm_2((\omega_1\omega_2\omega_3))=1, \\
&PMF_3(\Omega): Perm_3(F_7^5)=1\ \ \Rightarrow\ Perm_3((\omega_3\omega_1\omega_2))=1. \notag
\end{aligned}
\end{equation}
Under the RFS interpretation, $F_3^1, F_7^1, F_7^5$ represent three distinct permutation events, each treated as a holistic observation. $PMF_1$ indicates the observation of two types of elements $\omega_1$ and $\omega_2$. Both $PMF_2$ and $PMF_3$ observe three types of elements $\omega_1$, $\omega_2$, and $\omega_3$.  Crucially, the elements $\omega_1, \omega_2$ commonly observed by $PMF_1$, $PMF_2$ and $PMF_3$ all appear in the order $\omega_1\omega_2$. Specifically, the presence of $\omega_3$ in $PMF_2$ (at the end) and $PMF_3$ (at the beginning) represents an equivalent deviation in terms of set cardinality and composition relative to $PMF_1$. The relative topological structure of the common elements remains invariant across all three observations. Consequently, the conflict between $RPS_1$ and $RPS_2$ arises solely from differences in observed element cardinality and type, a situation identical to that between $RPS_1$ and $RPS_3$. Thus, from the RFS perspective, we conclude that $Conf(RPS_1, RPS_2)=Conf(RPS_1, RPS_3)$.

Now, we are starting from another direction. Under the DST perspective, if there is currently no more information available for fusion, classic TBM distributes beliefs evenly to singletons based on the insufficient reason principle\cite{smets1994transferable}. When the order of elements is introduced under this interpretation, its meaning will differ from that in RFS. Order now serves as the propensity of the decision-maker's belief transfer, called qualitative propensity. For the PMF $Perm((\omega_1\omega_2))=t$, it represents that the belief with $t$ has a higher propensity to transfer to $\omega_1$. In the layer-2 belief structure proposed in \cite{zhou2023conjunctive}, the order is a symbolic constraint, which only provides the direction of belief propensity transfer at the decision level, but it is not enough to affect or change the numerical value of the belief at the credal level.

We still observe the three RPSs in Example 3.1. From the perspective of DST, $PMF_1$ describes the target element as $\omega_1\succ\omega_2$. The decision-maker believes the observed target is $\omega_1$ or $\omega_2$. When there is no more information input, and a decision must be made, the decision-maker prefers $\omega_1$. Both $PMF_2$ and $PMF_3$ show that the possible target elements are $\omega_1, \omega_2$, and $\omega_3$. But the propensity displayed in $PMF_2$ is $\omega_1\succ\omega_2\succ\omega_3$, and in $PMF_3$ is $\omega_3\succ\omega_1\succ\omega_2$. It is obvious that the direction of belief transfer tendency of $PMF_1$ and $PMF_2$ is $\omega_1$, followed by $\omega_2$, while $PMF_3$ prefers $\omega_3$. Therefore, the conflict between $RPS_1$ and $RPS_2$ simply because of the unobserved target $\omega_3$ is much smaller than the conflict between $RPS_1$ and $RPS_3$ due to the inconsistent qualitative propensity of the target element. From the perspective of DST, there is $Conf(RPS_1, RPS_2)< Conf(RPS_1, RPS_3)$

In this paper, we focus on the conflict measure between RPSs following the latter perspective, treating RPST as an extension of the DST view. Under this interpretation, an essential characteristic of ordered focal sets is that higher-ranked elements occupy a more critical position.

\subsection{Discussion on the existing conflict measure in RPS}
The conflict measure between mass functions is a proper extension of conflict measures between sets. Similarly, the properties of conflict measure between permutation mass functions should be based on properties that appear natural in measuring conflict between permutations. This means that when the permutation mass function reduces to permutations, the conflict measure should also reduce to the one used between permutations. Therefore, permutations are here our starting point for defining the ideal properties of a conflict measure between permutation mass functions.

For two sets $F_i$ and $F_j$, we usually determine the conflict based on whether the intersection between them is empty, which can be expressed as
\begin{equation}\label{eq19}
    k_{set}(F_i, F_j)=\begin{cases}
          1   & F_i\cap F_j=\emptyset \\
          0   & F_i\cap F_j\neq\emptyset
    \end{cases}
\end{equation}
Moreover, Destercke and Burger formulate some desirable properties of a conflicting measure between sets in \cite{destercke2013toward}.

$\mathbf{Example \ 3.2.}$ Let $m_1$ and $m_2$ be two BBAs from different sources on the frame $\Omega=\{\omega_1, \omega_2, \omega_3\}$ respectively,
\begin{equation}
\begin{aligned}
&m_1(\Omega): m_1(F_3)=0.8, \ \ \ \ m_1(F_4)=0.2;   \\
&\ \ \ \Rightarrow  \ \ \   m_1(\{\omega_1, \omega_2\})=0.8, \ \ \ \    m_1(\{\omega_3\})=0.2; \\
&m_2(\Omega): m_2(F_7)=1.                      \\
&\ \ \ \Rightarrow  \ \ \   m_2(\{\omega_1, \omega_2, \omega_3\})=1.      \notag
\end{aligned}
\end{equation}
It is obvious that there is no conflict between the focal set $\mathcal{F}_1$ in $m_1$, and the focal set $\mathcal{F}_2$ in $m_2$, since $F_3\cap F_7\neq\emptyset$, $F_4\cap F_7\neq\emptyset$. According to formula (\ref{eq3}) in Definition 2.3, the conflict between $m_1$ and $m_2$ is calculated as $k=0$.

However, after introducing order information into sets, the situation becomes different. The judgment condition of whether the intersection is empty becomes invalid when dealing with permutations. We further elaborate through the following example, which only adds order information in Example 3.2.

$\mathbf{Example \ 3.3.}$ Let us consider $RPS_1$ and $RPS_2$ from two distinct sources on the frame $\Omega=\{\omega_1, \omega_2, \omega_3\}$ , and their PFMs are as follows
\begin{equation}
\begin{aligned}
&PMF_1(\Omega): Perm_1(F_3^1)=0.8, \ \ \ \ Perm_1(F_4^1)=0.2;   \\
&\ \ \ \ \ \ \Rightarrow  \ \ \ \ \  Perm_1((\omega_1\omega_2))=0.8, \ \ \ \    Perm_1((\omega_3))=0.2; \\
&PMF_2(\Omega): Perm_2(F_7^4)=1.                      \\
&\ \ \ \ \ \ \Rightarrow  \ \ \ \ \  Perm_2((\omega_2\omega_3\omega_1))=1.       \notag
\end{aligned}
\end{equation}
According to formulas (\ref{eq9}) and (\ref{eq10}), there are
\begin{equation}
\begin{aligned}
&F_3^1\overset{\leftarrow}{\cap}F_7^4=(\omega_1\omega_2), \ \ \ \ F_3^1\overset{\rightarrow}{\cap}F_7 ^4=(\omega_2\omega_1);   \\
&F_4^1\overset{\leftarrow}{\cap}F_7^4=(\omega_3), \ \ \ \ \ \ \ F_4^1\overset{\rightarrow}{\cap}F_7 ^4=(\omega_3).   \notag
\end{aligned}
\end{equation}
Finally, using formulas (\ref{eq7}) and (\ref{eq8}), the left and right conflict between $Perm_1$ and $Perm_2$ are respectively calculated as $\overset{\leftarrow}{K}=0$ and $\overset{\rightarrow}{K}=0$.

It can be seen that the conflict measure of permutation mass functions proposed in the original RPS literature\cite{deng2022random} still handles conflict between permutations in the same way as it deals with sets. The conflict is judged by whether the permutation intersection is empty. However, the left and right intersection of permutations focuses more on the order in which the intersection result obeys one party. The condition used to judge the logical consistency between sets, whether the intersection is empty, cannot effectively reflect the order consistency between the permutations. Although the intersection between the ordered focal sets $\mathcal{OF}_1$ in $PMF_1$ and $\mathcal{OF}_2$ in $PMF_2$ is non-empty, there is still an order conflict between them. $PMF_1$ conveys information about the target element with a belief of 0.8 as $\omega_1\succ\omega_2$ and with a belief of 0.2 as $\omega_3$. $PMF_2$ shows that the belief with 1 has transfer propensity between target elements is $\omega_2\succ\omega_3\succ\omega_1$. The order inconsistency here reflects the inconsistency in belief transfer propensity. Therefore, it is inappropriate to simply conclude that there is no conflict between $RPS_1$ and $RPS_2$.

All in all, there is an urgent need for a conflict measure method that can measure the order conflict between permutations and satisfy the viewpoint from the DST perspective, which is also the basis for measuring conflicts between PMFs.

\subsection{The proposed conflict measure method}
In this section, we define conflict between permutations, that is, the degree of inconsistency between permutations, which is the inverse of the consistency degree. Furthermore, a conflict measure method between permutation mass functions in RPS is proposed.

Suppose there are two permutations $A$ and $B$. $A_i$ is the $i$th element of permutation $A$. $A_{l_1:l_2}=\{A_i:l_1\le i\le l_2\}$ is a permutation consist of all elements from the $l_1$-th to $l_2$-th positions in the permutation. According to formula (\ref{eq13}), when the depth is $d$, the overlap of $A$ and $B$ is $|I_d|=|A_{1:d}\cap B_{1:d}|$. The non-overlap of $A$ and $B$ at depth $d$ is $max(|A_{1:d}|, |B_{1:d}|)-|I_d|$. Therefore, the conflict ratio, or degree of inconsistency, between permutations $A$ and $B$ at depth $d$ can be defined as
\begin{equation}
	\label{eq20}
	InC_d=\frac{max(|A_{1:d}|, |B_{1:d}|)-|A_{1:d}\cap B_{1:d}|}{d}.
\end{equation}

$\mathbf{Definition \ 3.1.}$ \textit{(Conflict of permutations)} For each $d\in\{1, 2, ..., n\}$, calculate the conflict ratio at $d$. A conflict between two permutations is defined as
\begin{equation}
    \label{eq21}
k_{perm}(A,B)=\sum_{d=1}^{n}\alpha_d\cdot InC_d.
\end{equation}
Where $\alpha_d$ is the weight when the depth is $d$. The choice of weights should always be guided by practical considerations relevant to the specific application. In this paper, we mainly focus on the following two cases:

(1) $\alpha_d=\frac{1}{n}$ for $d=1, 2, ..., n$, that is, each depth has the same weight.

(2) $\alpha_d=(1-p)\cdot p^{d-1}$, $p\in(0, 1)$. When $0< p< 1$ and $d$ approaches infinity, there is $\sum_{d=1}^{\infty}p^{d-1}=\frac{1}{1-p}$, then $ \sum_{d=1}^{\infty}(1-p)\cdot p^{d-1}=1$. In this case, $k_{perm}$ has the potential to measure infinite permutations. The parameter $p$ characterizes the reduction in weight. The smaller $p$ means the greater the weight at the top and the steeper the weight decline. $p=0$ indicates that only the front target element of permutations is considered. In contrast, $p=1$ indicates that the weights are arbitrarily flat but meaningless.

If there is no additional statement, the default $n=max(|A|,|B|)$. In particular, the decision-maker can truncate the permutation at any depth. At this time, $n$ takes any integer value from 1 to $max(|A|,|B|)$, which means that the decision-maker only pays attention to the first $n$ target elements.

According to the definition $C_d \in [0, 1]$, $InC_d \in [0, 1]$, it is easy to prove that the value of $k_{perm}$ also falls within the unit interval $[0, 1]$. The extreme value $k_{perm}$=1 represents two total conflicting permutations, and $k_{perm}$=0 represents consistent permutations.

$\mathbf{Definition \ 3.2.}$ \textit{(Conflict of permutation mass functions)} Given two RPSs defined on the frame $\Omega$ from two distinct sources, their permutation mass functions are $Perm_1$ and $Perm_2$, respectively. The conflict measure between $RPS_1$ and $RPS_2$ is defined as
\begin{equation}
    \label{eq22}
Conf(RPS_1, RPS_2)= \sum_{F_i^m\neq F_j^n}k_{perm}(F_i^m, F_j^n)\cdot Perm_1(F_i^m)Perm_2(F_j^n).
\end{equation}

Later, we will provide a detailed discussion on some properties of the proposed conflict measure. Now, we use the following numerical example to demonstrate the specific calculation process of the proposed method in detail.

$\mathbf{Example \ 3.4.}$ Let us consider $RPS_1$ and $RPS_2$ from two distinct sources on a frame with five elements, and their PFMs are as follows
\begin{equation}
\begin{aligned}
&PMF_1(\Omega): Perm_1(F_{23}^7)=0.6, \ \ \ \ Perm_1(F_{21}^2)=0.4;   \\
&\ \ \ \ \ \ \Rightarrow  \ \ \ \ \  Perm_1((\omega_2\omega_1\omega_3\omega_5))=0.6, \ \ \ \  Perm_1((\omega_1\omega_5\omega_3))=0.4.   \\
&PMF_2(\Omega): Perm_2(F_{31}^1)=0.8, \ \ \ \ Perm_1(F_{5}^1)=0.2;                       \\
&\ \ \ \ \ \ \Rightarrow  \ \ \ \ \   Perm_2((\omega_1\omega_2\omega_3\omega_4\omega_5))=0.8, \ \ \ \    Perm_2((\omega_1\omega_3))=0.2;     \notag
\end{aligned}
\end{equation}
The first step is calculating the conflict between the ordered focal sets $\mathcal{OF}_1$ in $PMF_1$ and $\mathcal{OF}_2$ in $PMF_2$. Using formula (\ref{eq20}), we list the conflict ratios between ordered focal sets $F_{23}^7$ and $F_{31}^1$ at each depth in Table \ref{tab2}.

 \begin{table}[!htbp]
    	\begin{center}
        	\caption{The conflict ratios between $F_{23}^7$ and $F_{31}^1$ at each depth in Example 3.4}
    		\begin{tabular}{cllcccc}
    			\toprule
    			$Depth$ &\ \ \ \ \ \ $F_{23}^7$ &\ \ \ \ \ \ \ \ $F_{31}^1$&$I_d$& $InC_d$ \\
    			\midrule
                \specialrule{0em}{1pt}{1pt}
    			1 & $\omega_2$ & $\omega_1$ & $\emptyset$	& 1 \\
    			\specialrule{0em}{1pt}{1pt}
    			2 & $\omega_2,\omega_1$ &$\omega_1,\omega_2$ &$\{\omega_1,\omega_2\}$ & 0 \\
        		\specialrule{0em}{1pt}{1pt}
    			3 &$\omega_2,\omega_1,\omega_3$  &$\omega_1,\omega_2,\omega_3$& $\{\omega_1,\omega_2,\omega_3\}$ & 0 \\
        		\specialrule{0em}{1pt}{1pt}
    			4  &$\omega_2,\omega_1,\omega_3,\omega_5$  &$\omega_1,\omega_2,\omega_3,\omega_4$ &$\{\omega_1,\omega_2,\omega_3\}$&$\frac{1}{4}$ \\
        		\specialrule{0em}{1pt}{1pt}
    			5  &$\omega_2,\omega_1,\omega_3,\omega_5$  &$\omega_1,\omega_2,\omega_3,\omega_4,\omega_5$& $\{\omega_1,\omega_2,\omega_3,\omega_5\}$ &$\frac{1}{5}$ \\
    			\bottomrule
    			\label{tab2}
    		\end{tabular}
    	\end{center}
    \end{table}

Then, formula (\ref{eq21}) is used to calculate $k_{perm}(F_{23}^7, F_{31}^1)$ as follows
$$k_{perm}(F_{23}^7, F_{31}^1)=\frac{1}{5}\cdot(1+0+0+\frac{1}{4}+\frac{1}{5})=0.29.$$
For simplicity, we directly take the average here, that is $\alpha_d=\frac{1}{n}$. Similarly, we have
$$k_{perm}(F_{23}^7, F_{5}^1)=0.5833, \ \ k_{perm}(F_{21}^2, F_{31}^1)=0.3467, \ \  k_{perm}(F_{21}^2, F_{5}^1)=0.2778. $$

Finally, according to formula (\ref{eq22}), we can get the conflict between $RPS_1$ and $RPS_2$ as
\begin{equation}
\begin{aligned}
Conf(RPS_1, RPS_2)&=0.29\times0.6\times0.8+0.5833\times0.6\times0.2+\\
&\ \ \ \ \  0.3467\times0.4\times0.8+0.2788\times0.4\times0.2   \\
&=0.3424       \notag
\end{aligned}
\end{equation}

If we use the existing method proposed in the original RPS literature\cite{deng2022random} to calculate the conflict between $Perm_1$ and $Perm_2$, due to $F_{23}^7\cap F_{31}^1\neq\emptyset$,  $F_{23}^7\cap F_{5}^1\neq\emptyset$,  $F_{21}^2\cap F_{31}^1\neq\emptyset$, and $F_{21}^2\cap F_{5}^1\neq\emptyset$, the final result is $\overset{\leftarrow}{K}=\overset{\rightarrow}{K}=0$. Compared with the existing method, the proposed conflict measure can more effectively reflect conflicts between permutation mass functions caused by order inconsistency.

\section{Properties and behaviour of the proposed conflict measure}
\label{4}
This section mainly provides a detailed analysis of the properties and behavior of the proposed conflict measure. In this paper, conflict is defined without relying on (in)dependent assumptions that are not supported by evidence.

\subsection{Characterizing conflict and non-conflict}
In \cite{destercke2013toward}, the conflict of mass functions is described as total conflict and non-conflict. The non-conflict is divided into three categories according to different restrictions on the intersection of focal sets. However, conflicts between permutation mass functions arise not only from the empty intersection of ordered focal sets but also from inconsistent ordering. This inconsistency can be caused by the same elements being placed in a different order or missing parts. When there is an intersection between permutations, we first consider the conflict of permutations into the following two categories.

$\mathbf{Definition \ 4.1.}$ \textit{(Order Conflict):} Given two permutations $F_i^m$ and $F_j^n$, which correspond to the index functions $\rho_1(\omega)$ and $\rho_2(\omega)$ respectively. The index function indicates the position of an element in the permutation. For example, for the permutation $\omega_3\omega_1\omega_2$, $\rho(\omega_1)=2, \rho(\omega_2)=3, \rho(\omega_3)=1$. There is an order conflict between permutations $F_i^m$ and $F_j^n$ iff $\exists$ $\omega\in F_i^m\cap F_j^n$ such that $\rho_1(\omega)\neq\rho_2(\omega)$.

$\mathbf{Definition \ 4.2.}$ \textit{(Elemental Conflict):} Permutations $F_i^m$ and $F_j^n$ are elemental conflict iff $F_i^m\cap F_j^n\neq\emptyset$, $F_i^m\overset{\leftarrow}{\cap} F_j^n= F_i^m\overset{\rightarrow}{\cap} F_j^n$, and $F_i^m\neq F_j^n$. Element conflict can also be defined as $\exists$ a set $\gamma$ such that $F_i^m\backslash\backslash\gamma=F_j^n$ or $F_j^n\backslash\backslash\gamma=F_i^m$, and $\gamma\neq\emptyset$.

$\mathbf{Example \ 4.1.}$ Assume that there are four permutations as follows, and the corresponding index functions are $\rho_1$, $\rho_2$, $\rho_3$, and $\rho_4$.
\begin{equation}
F_1^1:\omega_1,\ \ \ \ F_2^1:\omega_2,\ \ \ \ F_3^1:\omega_1\omega_2,\ \ \ \ F_3^2:\omega_2\omega_1. \notag
\end{equation}
Using the formula (\ref{eq21}) to calculate the conflict between them, we can get
\begin{equation}
\begin{aligned}
&k_{perm}(F_1^1, F_2^1)=1, \ \ \ k_{perm}(F_1^1, F_3^1)=0.25, \ \ \ k_{perm}(F_1^1, F_3^2)=0.75 \\
&k_{perm}(F_2^1, F_3^1)=0.75, \ \ \ k_{perm}(F_2^1, F_3^2)=0.25,\ \ \ k_{perm}(F_3^1, F_3^2)=0.5.   \notag
\end{aligned}
\end{equation}

There is no doubt a total conflict between $F_1^1$ and $F_2^1$ because $F_1^1\cap F_2^1=\emptyset$. Therefore, $k_{perm}(F_1^1, F_2^1)$ takes the extreme value 1. In particular, when the intersection between two permutations is non-empty, e.g., $F_1^1\cap F_3^1$, $F_1^1\cap F_3^2$, $F_2^1\cap F_3^1$, $F_2^1\cap F_3^2$, $F_3^1\cap F_3^2$, there is still $k_{perm}>0$, which indicates a partial conflict. According to Definitions 4.1 and 4.2, there is only element conflict between $F_1^1$ and $F_3^1$, since $\rho_1(\omega_1)=\rho_3(\omega_1)$, and $F_3^1\backslash\backslash\omega_2=F_1^1$. The same goes for permutations $F_2^1$ and $F_3^2$. Conflicts between $F_3^1$ and $F_3^2$ for permutations with identical elements are caused only by different order of elements. For the permutations $F_1^1$ and $F_3^2$ (same for $F_2^1$ and $F_3^1$), there are both element and order conflicts. An interesting finding is $k_{perm}(F_1^1, F_3^2)=k_{perm}(F_2^1, F_3^2)+k_{perm}(F_3^1, F_3^2)$, which seems to imply that such conflicts are linearly additive. There are also $k_{perm}(F_2^1, F_3^1)=k_{perm}(F_1^1, F_3^1)+k_{perm}(F_3^1, F_3^2)$. When we further generalize to the 3-element permutations, we still have $k_{perm}(F_7^1, F_4^1)=k_{perm}(F_7^1, F_1^1)+k_{perm}(F_7^1, F_7^6)$.

Next, consider two RPSs defined on the same frame from different sources, with PMFs $Prem_1$ and $Prem_2$, respectively. Based on the order conflict and elemental conflict in the ordered focal sets, we divide conflicts into four levels, from strongest to weakest, and deploy specific examples of different cases in Table \ref{tab3}.

$\mathbf{Definition \ 4.3.}$ \textit{(Total Conflict):} Permutation mass functions $Prem_1$ and $Prem_2$ are total conflicting when $\mathcal{OF}_1\cap \mathcal{OF}_2=\emptyset$.

$\mathbf{Definition \ 4.4.}$ \textit{(Strong Conflict):} Permutation mass functions $Prem_1$ and $Prem_2$ are strong conflict iff $\forall$ $F_i^m$, $F_j^n$ such that $F_i^m\in\mathcal{OF}_1$, $F_j^n\in\mathcal{OF}_2$, we have both order conflict and element conflict between $F_i^m$ and $F_j^n$.

$\mathbf{Definition \ 4.5.}$ \textit{(Weak Conflict):} Permutation mass functions $Prem_1$ and $Prem_2$ are weak conflict iff $\forall$ $F_i^m$, $F_j^n$ such that $F_i^m\in\mathcal{OF}_1$, $F_j^n\in\mathcal{OF}_2$, we have order conflict or element conflict between $F_i^m$ and $F_j^n$.

$\mathbf{Definition \ 4.6.}$ \textit{(Non-Conflict):} Permutation mass functions $Prem_1$ and $Prem_2$ are non-conflict iff
$$\underset{F_o^l\in\{\mathcal{OF}_1\cup\mathcal{OF}_2\}}{\bigcap}F_o^l=F_o^l$$

Total conflict indicates that the focus elements of the two sources have no intersection at all, such as Case 1 and Case 2 in Table \ref{tab3}. If $Perm_1$ and $Perm_2$ are categorical, there is only one ordered focal set $F_i^j\in OF$ such that $Perm(F_i^j)=1$, such as Case 1, Definition 4.3 reduces to the empty intersection of the permutations. Cases 3 and Cases 4 are strong conflicts that require both order and element conflicts between ordered focal sets $\mathcal{OF}_1$ and $\mathcal{OF}_2$. Here, we treat the empty intersection set as a stronger conflict than order and element conflicts. If there is only one type of conflict, order conflict or element conflict, between PMFs, it can be judged as a weak conflict. For example, $PMF_1$ and $PMF_2$ in Cases 5 and Cases 6 only have order conflicts, while Cases 7 and Cases 8 only have element conflicts. Only when two PMFs are identical and categorical can we determine that they are non-conflict, as shown in Case 9.

 \begin{table}[!htbp]
 \small
    	\begin{center}
    		\caption{Nine cases of two PMFs on a frame with seven elements. A pair of values $<(\omega_1\omega_2\omega_3), 1>$ in Case 1 under column name $RPS_1$ means $Perm_1(\omega_1\omega_2\omega_3)=1$.}
    		\begin{tabular}{cllccc}
    			\toprule
    			Case &  $RPS_1:$ &$RPS_2:$  &  $Conf$  \\
    			\midrule
    			\specialrule{0em}{1pt}{1pt}
    			1 & $\{<(\omega_1\omega_2\omega_3), 1>\}$ &$\{<(\omega_4\omega_5\omega_6), 1>\}$  &1.0000  \\
        		\specialrule{0em}{1pt}{1pt}
    			2 & $\{<(\omega_1\omega_2), 0.8>, <(\omega_4\omega_7), 0.2>\}$ &$\{<(\omega_3\omega_5), 0.5>, <(\omega_6), 0.5>\}$  &1.0000 \\
        		\specialrule{0em}{1pt}{1pt}
    			3  & $\{<(\omega_3\omega_5\omega_7), 1>\}$ &$\{<(\omega_5\omega_7), 0.4>, <(\omega_7\omega_3), 0.6>\}$  &0.6111    \\
        		\specialrule{0em}{1pt}{1pt}
    			4  & $\{<(\omega_1\omega_2), 0.3>, <(\omega_3\omega_4), 0.7>\}$ &$\{<(\omega_2\omega_3\omega_4), 0.5>, <(\omega_5), 0.5>\}$  &0.8222  \\
                \specialrule{0em}{1pt}{1pt}
    			5 &  $\{<(\omega_2\omega_3\omega_4), 1>\}$ &$\{<(\omega_2\omega_4\omega_3), 1>\}$  &0.1667   \\
                \specialrule{0em}{1pt}{1pt}
    			6 &  $\{<(\omega_2\omega_3\omega_4), 1>\}$ &$\{<(\omega_3\omega_2\omega_4), 0.5>, <(\omega_4\omega_2\omega_3), 0.5>\}$  &0.4167   \\
                \specialrule{0em}{1pt}{1pt}
    			7 &  $\{<(\omega_2\omega_3\omega_4), 1>\}$ &$\{<(\omega_2\omega_3), 1>\}$  &0.1111   \\
                \specialrule{0em}{1pt}{1pt}
    			8 &  $\{<(\omega_2\omega_3\omega_4), 1>\}$ &$\{<(\omega_2), 0.8>, <(\omega_2\omega_3), 0.2>\}$  &0.3333   \\
                \specialrule{0em}{1pt}{1pt}
    			9 &  $\{<(\omega_1\omega_7\omega_2\omega_5), 1>\}$ &$\{<(\omega_1\omega_7\omega_2\omega_5), 1>\}$  &0.0000   \\
    			\bottomrule
    			\label{tab3}
    		\end{tabular}
    	\end{center}
    \end{table}

\subsection{Properties of the proposed conflict measure}
We formulate the properties of conflict measure $Conf: PMF_{\Omega}\times PMF_{\Omega} \rightarrow  [0, 1]$ and elaborate on the properties through some numerical examples.

$\mathbf{Property \ 4.1.}$ \textit{(Extreme Conflict Values):} $Conf(RPS_1, RPS_2)=1$, if and only if $Perm_1$ and $Perm_2$ are in total conflict. $Conf(RPS_1, RPS_2)=0$, if and only if $Perm_1$ and $Perm_2$ are non-conflicting.

In Table \ref{tab3} of the previous section, we have given examples of these two types of extreme conflict values, corresponding to Case 1 and Case 9, respectively.

$\mathbf{Property \ 4.2.}$ \textit{(Symmetry):} $Conf(RPS_1, RPS_2)=Conf(RPS_2, RPS_1)$

According to the definition $k_{perm}$ is symmetric, that is, $k_{perm}(A, B)=k_{perm}(B, A)$, so it is not difficult to prove that $Conf(RPS_1, RPS_2)=Conf(RPS_2, RPS_1)$.

$\mathbf{Property \ 4.3.}$ \textit{(Top-weightiness):} The top-ranked elements in order focal sets have a more critical status. In other words, inconsistent ordering of higher-ranked elements will lead to more significant conflicts.

$\mathbf{Example \ 4.2.}$ Suppose there are $RPS_1$ and $RPS_2$ from two distinct sources defined on a frame with eight elements. Their PFMs are as follows.
\begin{equation}
\begin{aligned}
&PMF_1(\Omega): Perm_1(F_{32}^1)=0.2, \ \ \ \ Perm_1(F_{192}^1)=0.3, \ \ \ \ Perm_1(F_{31}^1)=0.5;   \\
&\ \ \ \ \ \ \Rightarrow  \ \ \ \ \  Perm_1((\omega_6))=0.2,\ Perm_1((\omega_7\omega_8))=0.3,\ Perm_1((\omega_1\omega_2\omega_3\omega_4\omega_5))=0.5.   \\
&PMF_2(\Omega): Perm_2((X))=1.                   \notag
\end{aligned}
\end{equation}
If the order of ordered focal sets is ignored, PMFs degenerate into mass functions in Dempster-Shafer theory.
\begin{equation}
\begin{aligned}
&  m_1(\Omega):  \  m_1(F_{32})=0.2, \ \ \ \ m_1(F_{192})=0.3, \ \ \ \ m_1(F_{31})=0.5;   \\
&  \ \  \Rightarrow  \ \ \  \ \  m_1(\{\omega_6\})=0.2,\ \ m_1(\{\omega_7, \omega_8\})=0.3,\ \ m_1(\{\omega_1, \omega_2, \omega_3, \omega_4, \omega_5\})=0.5   \\
&  m_2(\Omega): \   m_2(\{X\})=1;                   \notag
\end{aligned}
\end{equation}

\begin{figure}[H]
	\centering
	\centerline{\includegraphics[width=4in,height=3in]{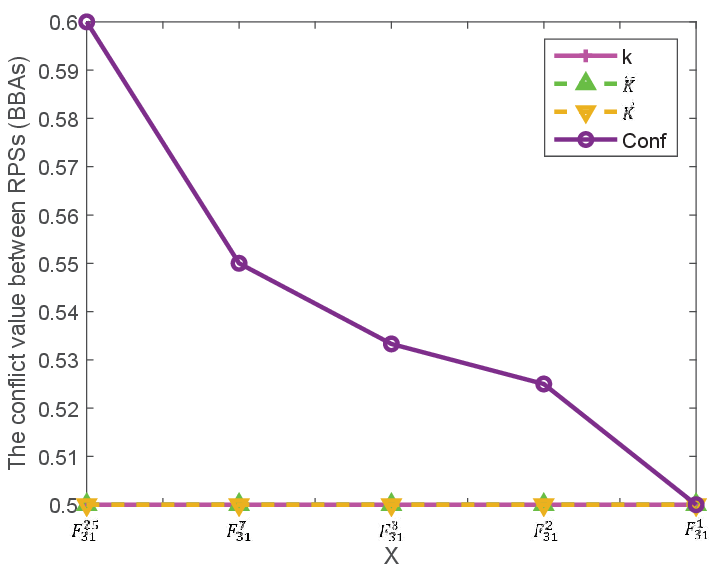}}
	\caption{The conflict between RPSs(BBAs) when $X$ takes different values in Example 4.2}
	\label{fig1}
\end{figure}

Table \ref{tab4} and Figure \ref{fig1} show the conflict between $RPS_1$ and $RPS_2$ calculated using the proposed method and the conflict measure in the original RPS literature\cite{deng2022random} when $X$ takes different values. Conflicts between the corresponding mass functions are also given. 

 \begin{table}[!htbp]
    	\begin{center}
    		\setlength{\tabcolsep}{0.5cm}
    		\caption{The conflict between RPSs(BBAs) when $X$ takes different values in Example 4.2}
    		\begin{tabular}{ccccccc}
    			\toprule
    			$X$ &  $k$\cite{shafer1976mathematical} &$\mathop{K}\limits^{\leftarrow}$\cite{deng2022random} & $\mathop{K}\limits^{\rightarrow}$\cite{deng2022random} & $Conf$  \\
    			\midrule
    			\specialrule{0em}{1pt}{1pt}
    			$F_{31}^{25}: \omega_2\omega_1\omega_3\omega_4\omega_5$ & 0.5000 &0.5000  &0.5000 &	0.6000  \\
        		\specialrule{0em}{1pt}{1pt}
    			$F_{31}^{7}: \omega_1\omega_3\omega_2\omega_4\omega_5$ &0.5000  &0.5000 &0.5000& 0.5500  \\
        		\specialrule{0em}{1pt}{1pt}
    			$F_{31}^{3}: \omega_1\omega_2\omega_4\omega_3\omega_5$  &0.5000  &0.5000 &0.5000 &  0.5333  \\
        		\specialrule{0em}{1pt}{1pt}
    			$F_{31}^{2}: \omega_1\omega_2\omega_3\omega_5\omega_4$  &0.5000  &0.5000 &0.5000&  0.5250  \\
                    \specialrule{0em}{1pt}{1pt}
    			$F_{31}^{1}: \omega_1\omega_2\omega_3\omega_4\omega_5$ & 0.5000 &0.5000 &0.5000 &	0.5000  \\
    			\bottomrule
    			\label{tab4}
    		\end{tabular}
    	\end{center}
    \end{table}
    
It can be found that $X$ is five different permutation events about element $\omega_1$, $\omega_2$, $\omega_3$, $\omega_4$, and $\omega_5$, with different orders. Since $X\cap F_{31}^1=\emptyset, X\cap F_{192}^1=\emptyset$, the change of ordered focal set $X$ can only affect the conflict between it and $F_{31}^1$, thereby affecting the conflict value between $RPS_1$ and $RPS_2$. Observe the conflict values obtained by the proposed conflict measure $Conf$. When the top two elements $\omega_1$ and $\omega_2$ are disrupted, the conflict between $RPS_1$ and $RPS_2$ is 0.6. When the order inconsistency appears in later and later elements, the conflict gradually becomes smaller until the last two elements are disrupted, and the conflict is 0.525. When $X$ is consistent with $F_{31}^1\in\mathcal{OF}_1$, the conflict reaches a minimum of 0.5.

For the corresponding mass function, the order of the elements is not considered, and $X$ degenerates into the focal set $F_{31}:\{\omega_1, \omega_2, \omega_3, \omega_4, \omega_5\}$. First, detect the conflict judgment conditions of the set, there are $F_{31}\cap F_{32}=\emptyset$, $F_{31}\cap F_{192}=\emptyset$, and $F_{31}\cap F_{31}\neq\emptyset$. According to the formula (\ref{eq3}), the conflict between the mass functions $m_1$ and $m_2$ is calculated as $k=0.2\times1+0.3\times1=0.5$.

In addition, the result obtained using the existing RPS conflict measure method\cite{deng2022random} is also $\mathop{K}\limits^{\leftarrow}=\mathop{K}\limits^{\rightarrow}=0.5$ because it specifies the same conditions for determining conflicts as the set in the processing mass function, i.e., whether the intersection is empty. Such judgment conditions make them believe there is no conflict between $F_{31}^1$ and $X$ because $F_{31}\cap X\neq\emptyset$, thus leading to unreasonable conflict values. In fact, there is still an order inconsistency between $F_{31}^1$ and $X$, which implies an inconsistency in the propensity of the decision-maker’s belief transfer.  The left and right intersection of permutation events only focuses on the obedience of the order after the intersection and cannot be used as a judgment condition for conflicts.

To sum up the above, the proposed RPS conflict measure method can effectively reflect the conflict between permutation mass functions. It has the natural top-weightiness property; the top-ranked elements are essential, and the inconsistent ordering of higher-ranked elements will lead to more significant conflicts, which is in line with the view of RPST from the perspective of DST.

$\mathbf{Example \ 4.3.}$ Assume there are $RPS_1$ and $RPS_2$ from two distinct sources defined on a frame with ten elements. Their PFMs are as follows.
\begin{equation}
\begin{aligned}
&PMF_1(\Omega): Perm_1(F_{8}^1)=0.05, \ \ \ \ Perm_1(F_{5}^1)=0.05, \ \ \ \ Perm_1((X))=0.8, \\
&\ \ \ \ \ \ \ \ \ \ \ \ \ \ \ \  Perm_1(F_{31}^1)=0.1;   \\
&\ \ \ \ \ \ \Rightarrow \ \ \ \ \  Perm_1((\omega_4))=0.05,\ \ Perm_1((\omega_2\omega_3))=0.05,\ \ Perm_1((X))=0.8,\\
& \ \ \ \ \ \ \  \ \ \ \ \ \ \ \ \  Perm_1((\omega_1\omega_2\omega_3\omega_4\omega_5))=0.1  \notag\\
&PMF_2(\Omega): Perm_2(F_7^1)=1.              \\
&\ \ \ \ \ \ \Rightarrow \ \ \ \ \ Perm_2((\omega_1\omega_2\omega_3))=1.  \notag
\end{aligned}
\end{equation}
If the order in ordered focal sets is not considered, PMFs degenerate into the following mass function.
\begin{equation}
\begin{aligned}
&m_1(\Omega): m_1(F_{8})=0.05, \  m_1(F_{5})=0.05, \ m_1(\{X\})=0.8,\ m_1(F_{31})=0.1;\\
& \ \ \ \Rightarrow \ \ \  m_1(\{\omega_4\})=0.05,\ \ m_1(\{\omega_2, \omega_3\})=0.05,\ \ m_1(\{X\})=0.8,\\
& \ \ \ \ \ \ \  \ \ \ \ m_1(\{\omega_1, \omega_2, \omega_3, \omega_4, \omega_5\})=0.1  \notag\\
&m_2(\Omega): m_2(F_7)=1.              \\
&\ \ \ \Rightarrow \ \ \  m_2(\{\omega_1, \omega_2, \omega_3\})=1.  \notag
\end{aligned}
\end{equation}

\begin{figure}[H]
	\centering
	\centerline{\includegraphics[width=4in,height=3in]{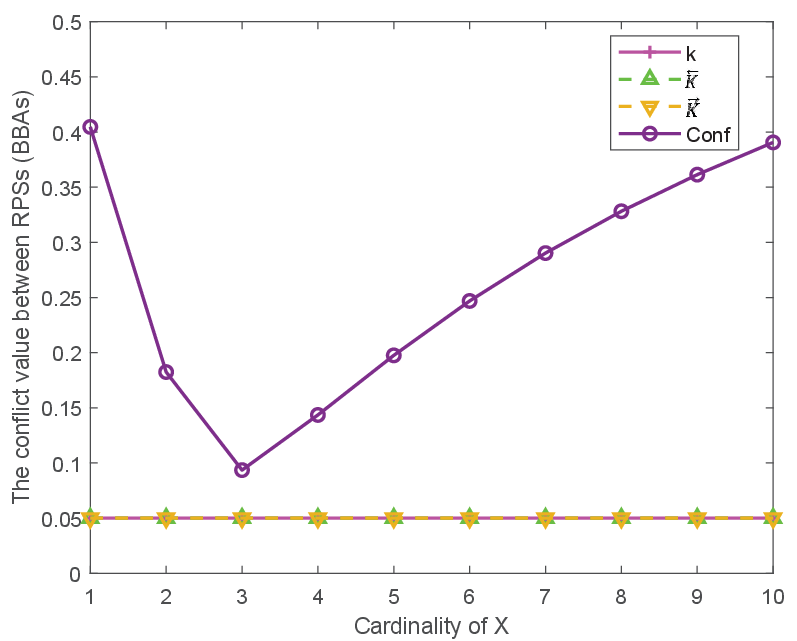}}
	\caption{The conflict between RPSs(BBAs) when $X$ changes in Example 4.3}
	\label{fig2}
\end{figure}

When $X$ changes from $\omega_1$ to $\omega_1\omega_2\omega_3...\omega_{10}$, the conflicts between RPSs and corresponding mass functions are as shown in Table \ref{tab5} and Figure \ref{fig2}.

\begin{table}[!htbp]
    	\begin{center}
        	\setlength{\tabcolsep}{0.5cm}
    		\caption{The conflict between RPSs (BBAs) when $X$ changes in Example 4.3}
    		\begin{tabular}{lcccccc}
    			\toprule
    			 \ \ \ \ \ \ \ \ \ \ \ \ \ \ \ \ \ $X$ &  $k$\cite{shafer1976mathematical}& $\mathop{K}\limits^{\leftarrow}$\cite{deng2022random} & $\mathop{K}\limits^{\rightarrow}$\cite{deng2022random} & $Conf$  \\
    			\midrule
    			\specialrule{0em}{1pt}{1pt}
    			$F_1^1: \omega_1$ & 0.0500 &0.0500  &0.0500 &	0.4047  \\
        		\specialrule{0em}{1pt}{1pt}
    			$F_3^1:\omega_1\omega_2$ &0.0500  &0.0500 &0.0500& 0.1824  \\
        		\specialrule{0em}{1pt}{1pt}
    			$F_7^1:\omega_1\omega_2\omega_3$  & 0.0500  &0.0500 &0.0500 &  0.0936  \\
        		\specialrule{0em}{1pt}{1pt}
    			$F_{15}^1:\omega_1\omega_2\omega_3\omega_4$  & 0.0500 &0.0500  &0.0500&  0.1436  \\
        		\specialrule{0em}{1pt}{1pt}
    			$F_{31}^1:\omega_1\omega_2\omega_3\omega_4\omega_5$  & 0.0500  &0.0500 &0.0500&  0.1976  \\
         		\specialrule{0em}{1pt}{1pt}
    			$F_{63}^1:\omega_1\omega_2\omega_3\omega_4\omega_5\omega_6$  & 0.0500 &0.0500  &0.0500&  0.2469  \\
         		\specialrule{0em}{1pt}{1pt}
    			$F_{127}^1:\omega_1\omega_2\omega_3\omega_4\omega_5\omega_6\omega_7$  & 0.0500  &0.0500 &0.0500& 0.2903   \\
         		\specialrule{0em}{1pt}{1pt}
    			$F_{255}^1:\omega_1\omega_2\omega_3\omega_4\omega_5\omega_6\omega_7\omega_8$  & 0.0500  &0.0500&0.0500&  0.3282   \\
         		\specialrule{0em}{1pt}{1pt}
    			$F_{511}^1:\omega_1\omega_2\omega_3\omega_4\omega_5\omega_6\omega_7\omega_8\omega_9$  & 0.0500  &0.0500&0.0500&  0.3614  \\
         		\specialrule{0em}{1pt}{1pt}
    			$F_{1023}^1:\omega_1\omega_2\omega_3\omega_4\omega_5\omega_6\omega_7\omega_8\omega_9\omega_{10}$  &0.0500 & 0.0500  &0.0500 & 0.3906   \\
    			\bottomrule
    			\label{tab5}
    		\end{tabular}
    	\end{center}
    \end{table}

It can be observed that only the proposed conflict measure $Conf$ can distinguish the impact of different $X$ in $RPS_2$ on the conflict between $RPS_1$ and $RPS_2$. When the cardinality of $X$ is closer to 3, the consistency between $X$ and $F_7^1$ is higher, and the conflict value between $RPS_1$ and $RPS_2$ is smaller. When $|X|=3$, there is $X=F_7^1$, and the conflict reaches the minimum of 0.0936. On the contrary, when the cardinality of $X$ increases or decreases, away from 3, the consistency between $X$ and $F_7^1$ gradually decreases, and the conflict gradually becomes more significant. More specifically, for the situation where the element conflict between X and $F_7^1$ is caused by one unobserved element, the conflict when $|X|=2$ is larger than the conflict when $|X|=4$. Similarly, for element conflict caused by two unobserved elements, the conflict when $|X|=1$ is larger than the conflict when $|X|=5$ or even $|X|=7$. This exactly reflects the top-weightiness nature of the proposed conflict measure method, where inconsistencies between top-ranked elements will cause greater conflicts. 

For the corresponding mass function, since there are $F_8\cap F_7=\emptyset$, $F_5\cap F_7\neq\emptyset$, $X\cap F_7\neq\emptyset$, $F_{31}\cap F_7\neq\emptyset$ , the conflict between $m_1$ and $m_2$ is calculated as $k=0.05\times1=0.05$, and remains unchanged. The conflict values of $RPS_1$ and $RPS_2$ calculated using the existing RPS conflict measure method\cite{deng2022random} have also remained at $\mathop{K}\limits^{\leftarrow}=\mathop{K}\limits^{\rightarrow}=0.05$. It still proves that the conflict judgment conditions in existing methods are not enough to judge order conflicts that can only occur between two permutations, let alone the inconsistency in the propensity of belief transfer of the decision-makers.

$\mathbf{Property \ 4.4.}$ \textit{(Arbitrarily Truncated):} Conflicts between two RPSs at any depth are allowed to be obtained. That is, we can truncate the ordered focal sets in RPS at any depth $n$, $n\in[1, max(|\mathcal{OF}_1|, |\mathcal{OF}_2|)]$, which means that the decision-maker only focuses on the first $n$ target elements and ignores the subsequent redundancy.

$\mathbf{Example \ 4.4.}$ Let us consider $RPS_1$ and $RPS_2$ from two distinct sources on the frame $\Omega=\{\omega_1, \omega_2, \omega_3, \omega_4, \omega_5, \omega_6, \omega_7\}$ , and their PFMs are as follows
\begin{equation}
\begin{aligned}
&PMF_1(\Omega): Perm_1(F_{31}^{31})=0.4, \ \ \ \ Perm_1(F_{6}^1)=0.6;  \\
&\ \ \ \ \ \ \Rightarrow \ \ \ \ \  Perm_1((\omega_2\omega_3\omega_1\omega_4\omega_5))=0.4,\ \ Perm_1((\omega_2\omega_3))=0.6;\\
&PMF_2(\Omega): Perm_2(F_{127}^{841})=1.              \\
&\ \ \ \ \ \ \Rightarrow \ \ \ \ \ Perm_2((\omega_2\omega_3\omega_1\omega_4\omega_5\omega_6\omega_7))=1.  \notag
\end{aligned}
\end{equation}
If the order in ordered focal sets is not considered, the mass function corresponding to PMFs is as follows.
\begin{equation}
\begin{aligned}
&m_1(\Omega): m_1(F_{31})=0.4, \  m_1(F_{6})=0.6;\\
&\ \ \ \Rightarrow \ \ \  m_1(\{\omega_1, \omega_2, \omega_3, \omega_4, \omega_5\})=0.4\ \ m_1(\{\omega_2, \omega_3\})=0.6. \\
&m_2(\Omega): m_2(F_{127})=1.              \\
&\ \ \ \Rightarrow \ \ \  m_2(\{\omega_1, \omega_2, \omega_3, \omega_4, \omega_5, \omega_6, \omega_7\})=1.  \notag
\end{aligned}
\end{equation}

\begin{figure}[H]
	\centering
	\centerline{\includegraphics[width=3.8in,height=3in]{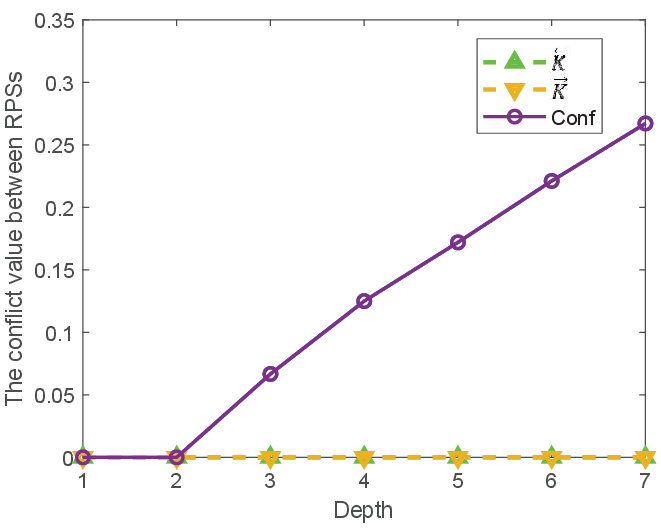}}
	\caption{The conflict between RPSs at different depths in Example 4.4}
	\label{fig3}
\end{figure}

The conflicts between $RPS_1$ and $RPS_2$ at different depths are calculated using the proposed and existing conflict measure methods\cite{deng2022random}, as shown in Table \ref{tab6} and Figure \ref{fig3}. Remarkably, in the mass function, the focal set is unordered, meaning the target elements are equally important and treated equally. Therefore, it makes no sense to consider the depth of the focus element; $k$ becomes inapplicable at this point.

\begin{table}[!htbp]
    	\begin{center}
        	\setlength{\tabcolsep}{0.5cm}
    		\caption{The conflict between RPSs at different depths in Example 4.4}
    		\begin{tabular}{lcccccc}
    			\toprule
    			$Depth$ &  $k$\cite{shafer1976mathematical}& $\mathop{K}\limits^{\leftarrow}$\cite{deng2022random} & $\mathop{K}\limits^{\rightarrow}$\cite{deng2022random} & $Conf$  \\
    			\midrule
    			\specialrule{0em}{1pt}{1pt}
    			$d=1$ & - &0.0000  &0.0000 &	0.0000  \\
        		\specialrule{0em}{1pt}{1pt}
    			$d=2$ &-  &0.0000 &0.0000& 0.0000  \\
        		\specialrule{0em}{1pt}{1pt}
    			$d=3$  & -  &0.0000 &0.0000 &  0.0667  \\
        		\specialrule{0em}{1pt}{1pt}
    			$d=4$ & - &0.0000  &0.0000&  0.1250  \\
        		\specialrule{0em}{1pt}{1pt}
    			$d=5$ & -  &0.0000 &0.0000&  0.1720  \\
         		\specialrule{0em}{1pt}{1pt}
    			$d=6$ & - &0.0000  &0.0000&  0.2211  \\
         		\specialrule{0em}{1pt}{1pt}
    			$d=7$ & - &0.0000 &0.0000& 0.2671   \\
    			\bottomrule
    			\label{tab6}
    		\end{tabular}
    	\end{center}
    \end{table}

By observing Table \ref{tab6} and Figure \ref{fig3}, we can find that the proposed method has the ability to measure conflicts between RPSs that are arbitrarily truncated.
$PMF_1$ shows that the belief with 0.4 has a transfer propensity between target elements is $\omega_2\succ\omega_3\succ\omega_1\succ\omega_4\succ\omega_5$, and with a belief of 0.6 as $\omega_2\succ \omega_3$. $PMF_2$ indicates that the target elements have transfer propensity $\omega_2\succ\omega_3\succ\omega_1\succ\omega_4\succ\omega_5\succ\omega_6\succ\omega_7$ with a belief of 1. When only focusing on the top two target elements, there is no conflict since $RPS_1$ and $RPS_2$ show the same qualitative propensity $\omega_2\succ\omega _3$. When focusing on the top three to top five target elements, $F_{31}^{31}$ and $F_{127}^{841}$ still have the same belief transfer propensity, while the conflict between $F_{6}^{1}$ and $F_{127}^{841}$ appears. Hence, the conflict between $RPS_1$ and $RPS_2$ gradually increases. When the depth reaches 6 or 7, that is, the first six or all target elements are to be focused, the inconsistency in $F_{31}^{31}$ and $F_{127}^{841}$ begins to appear, the conflict between $RPS_1 $ and $RPS_2$ further intensified.
However, since at any depth, there are $F_{31}^{31}\cap F_{127}^{841}\neq\emptyset$, $F_{6}^{1}\cap F_{127}^ {841}\neq\emptyset$, so using the existing RPS conflict measure method\cite{deng2022random} yields unreasonable results $\mathop{K}\limits^{\leftarrow}=\mathop{K}\limits^{\rightarrow}=0$, which cannot reflect the conflict between $RPS_1$ and $RPS_2$. Generally, as depth increases, the risk of inconsistencies increases. The proposed conflict measure method provides decision-makers with an option where we can focus only on what we want to focus on.

In the above, we discussed the situation with the same weight at each depth, that is, $\alpha_d=\frac{1}{n}$, $d=1, 2,..., n$. Next, we will analyze the unique properties of the proposed conflict measure if $\alpha_d=(1-p)\cdot p^{d-1}, p\in(0, 1)$ is selected as the weight.

$\mathbf{Property \ 4.5.}$ \textit{(Infinite Depth):} $Conf$ has the potential to measure conflicts between RPSs with infinite depth.

When $p\in(0, 1)$ and $d\rightarrow\infty$, there is $\sum_{d=1}^{\infty}p^{d-1}=\frac{1}{1-p}$, therefore $\sum_{d=1}^{\infty}\alpha_d=\sum_{d=1}^{\infty}(1-p)\cdot p^{d-1}=1$. For this reason, $k_{perm}$ has the ability to measure infinite permutations, which allows $Conf$ to measure conflicts between RPSs of infinite depth.

\begin{figure}[H]
	\centering
	\centerline{\includegraphics[width=4.8in,height=3in]{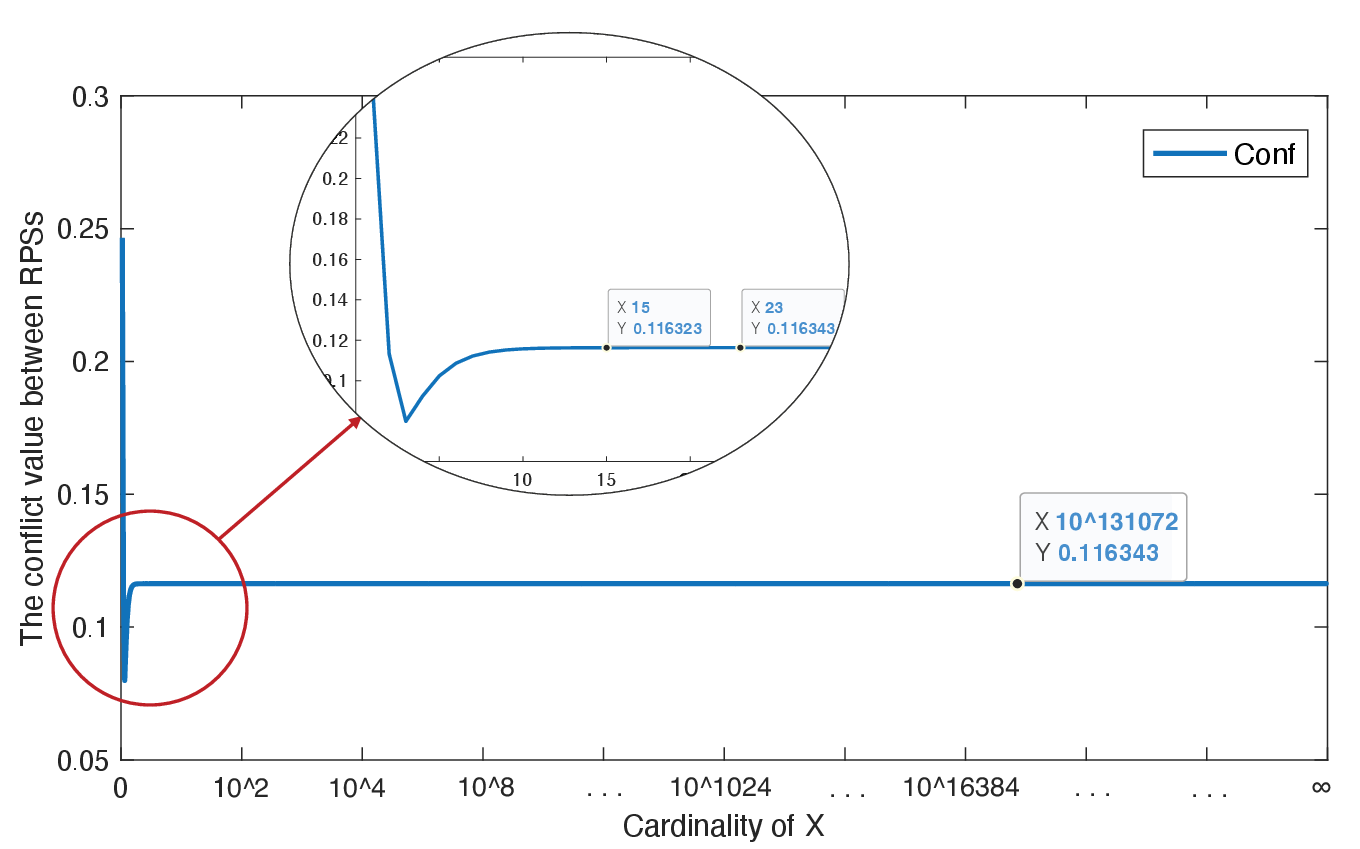}}
	\caption{The conflict between RPSs when $|X|$ from 1 to $\infty$ in Example 4.3}
	\label{fig4}
\end{figure}

Let us consider extending the cardinality of $X$ in Example 4.3 to infinity. Assuming that the sensor continuously observes and updates information, the change of $X$ indicates that new and less essential targets are continually monitored. Figure \ref{fig4} shows the conflicts between RPSs with infinite depth obtained using the proposed conflict measure method. For the convenience of the display, here we only select $p=0.5$. Later, we will specifically discuss the impact of different choices of $p$ on conflict.

The conflict between $RPS_1$ and $RPS_2$ still reaches the minimum at $|X|=3$. As $|X|$ gets further away from 3, the conflict continues to increase. When $3<|X|<10$, there is an obvious upward curve, although the growth rate is gradually declining. When $10<|X|<23$, the curve becomes flat, but a slight numerical increase can still be observed. After $|X|>23$, the curve appears to be a straight line parallel to the abscissa axis. This is because the larger the cardinality of $X$, the deeper the depth reached when calculating conflicts. As the depth $d$ increases, the weight of each depth is continuously weakened by a factor of 0.5, which directly weakens the impact of inconsistencies caused by lower-ranked elements on conflicts. In fact, the conflict is still increasing at a magnitude that is difficult to observe with the naked eye. Until at a certain depth $d$, the weight $\alpha_d$ approaches 0 infinitely. The inconsistent behavior of the target elements ranked after the $d$th position in the ordered focal set has no contribution to the conflict, and the conflict value converges.

\begin{figure}[H]
	\centering
	\centerline{\includegraphics[width=3.8in,height=3in]{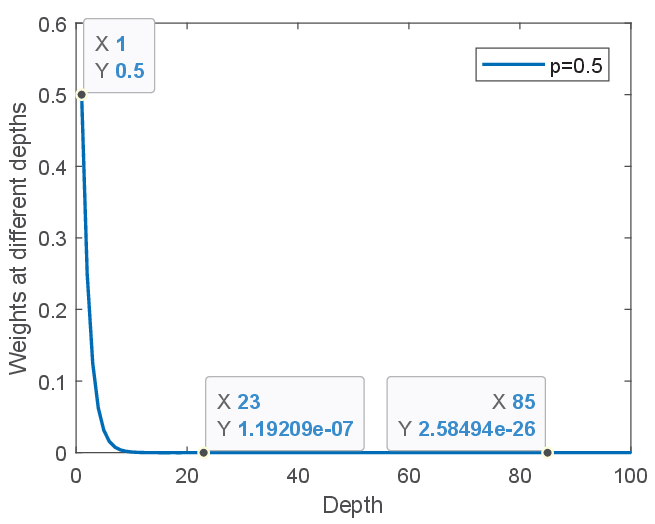}}
	\caption{Weights with depth from 1 to 100 when $p=0.5$}
	\label{fig5}
\end{figure}

In Figure \ref{fig5}, we show the weights corresponding to depths from 1 to 100 when $p=0.5$ so that we can more intuitively observe the decreasing trend of weights with depth. It can be found that when the depth $d$ reaches about 10, the value of the weight has been reduced from 0.5 to close to 0. When the depth $d=23$, the weight assigned is only $10^{-7}$ level, which is already a very small value, and its impact on the conflict between RPSs is almost negligible. Not to mention the situation with $d=85$ or deeper. The introduction of parameter $p$ not only enables the proposed overlap-based conflict measure method to have the ability to measure conflicts between RPSs with infinite depth but also makes the characteristics of top-weightiness more prominent. At this time, decision-makers can also take appropriate truncation measures to eliminate subsequent redundancies to improve decision-making efficiency.

$\mathbf{Property \ 4.6.}$ \textit{(Adjustable parameter):} An adjustable parameter $p\in (0, 1)$ is provided here so decision-makers can choose the appropriate $p$ based on actual application requirements.

\begin{figure}[H]
	\centering
	\centerline{\includegraphics[width=3.8in,height=3in]{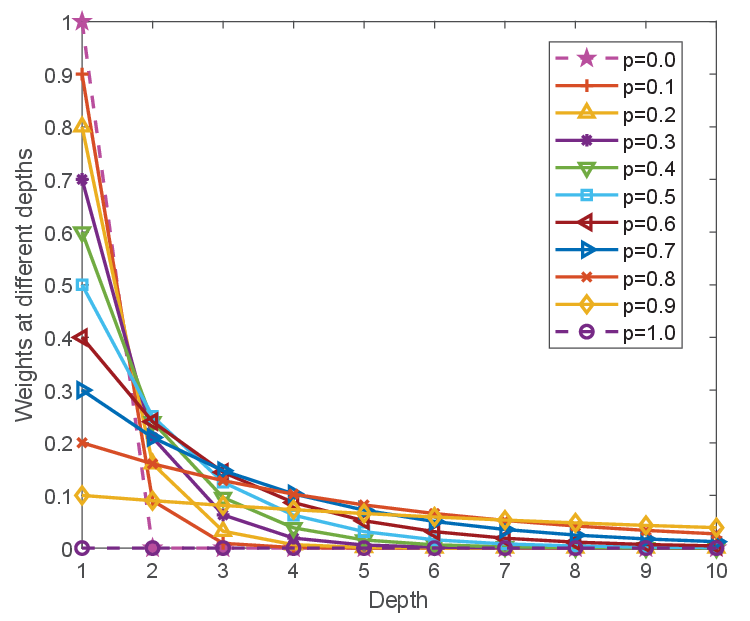}}
	\caption{The change of weights with depth when $p$ takes different values}
	\label{fig6}
\end{figure}

Figure \ref{fig6} provides the weights at depths 1 to 10 when $p=0, 0.1, 0.2,...,1$. The parameter $p$ characterizes how steep the weight decreases. A smaller $p$ indicates a greater weight at the top and a steeper drop in weight. When $p$ takes the limit value 0, only the top is assigned a weight of 1, which means that the decision-maker only cares about the first target elements in the ordered focal sets at this time. On the contrary, when $p$ takes 1, the weight is 0 at any depth, which is meaningless.

\begin{figure}[H]
	\centering
	\centerline{\includegraphics[width=3.8in,height=3in]{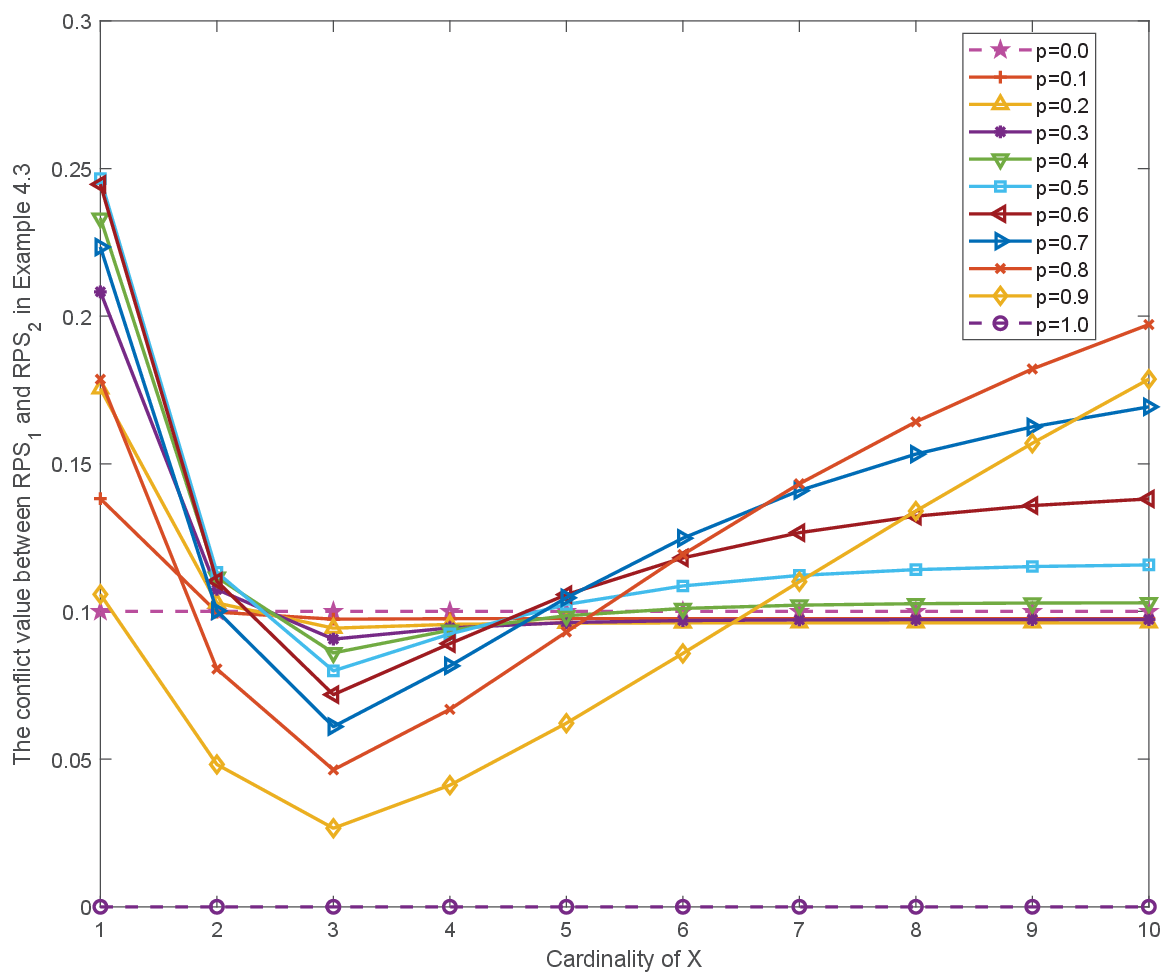}}
	\caption{The conflict between RPSs when $p$ takes different values in Example 4.3}
	\label{fig7}
\end{figure}

Here, we still use Example 4.3 to illustrate the impact of different values of parameter $p$ on conflicts between RPSs. In Table \ref{tab7} and Figure \ref{fig7}, we show the conflict values between $RPS_1$ and $RPS_2$ in Example 4.3 obtained by the proposed conflict measure method when $p=0, 0.1, 0.2, ...,1$. To display it more clearly, we only present the numerical results and curve graphs of conflicts with depths from 1 to 10 and the conflict convergence values.

\begin{table}[!htbp]\small
    	\begin{center}
        	\setlength{\tabcolsep}{0.5cm}
    		\caption{The conflict between RPSs when $p$ takes different values in Example 4.3}
    		\begin{tabular}{p{15em}p{2em}p{2em}p{2em}p{2em}p{2em}}
    			\toprule
                    &\multicolumn{5}{c}{$Conf$}  \\
                    \cmidrule(lr){2-6}
    			 \ \ \ \ \ \ \ \ \ \ \ \ \ \ \ \ \ $X$ & p=0.0 &p=0.1 & p=0.2 & p=0.3 &p=0.4\\
    			\midrule
    			\specialrule{0em}{1pt}{1pt}
    			$F_1^1: \omega_1$ & 0.1000 &0.1382  &0.1754 & 0.2082 &0.2332 \\
        		\specialrule{0em}{1pt}{1pt}
    			$F_3^1:\omega_1\omega_2$ & 0.1000 &0.0998 &0.1029& 0.1074 & 0.1116\\
        		\specialrule{0em}{1pt}{1pt}
    			$F_7^1:\omega_1\omega_2\omega_3$  &  0.1000  &0.0974 &0.0943 &  0.0906& 0.0860    \\
        		\specialrule{0em}{1pt}{1pt}
    			$F_{15}^1:\omega_1\omega_2\omega_3\omega_4$  & 0.1000&0.0976  &0.0956&  0.0944 &  0.0937    \\
        		\specialrule{0em}{1pt}{1pt}
    			$F_{31}^1:\omega_1\omega_2\omega_3\omega_4\omega_5$  & 0.1000 &0.0976  &0.0960&  0.0962& 0.0986   \\
         		\specialrule{0em}{1pt}{1pt}
    			$F_{63}^1:\omega_1\omega_2\omega_3\omega_4\omega_5\omega_6$  &  0.1000 &0.0976  &0.0961&  0.0969 & 0.1010    \\
         		\specialrule{0em}{1pt}{1pt}
    			$F_{127}^1:\omega_1\omega_2\omega_3\omega_4\omega_5\omega_6\omega_7$  &  0.1000  &0.0976  &0.0962& 0.0972 &  0.1022    \\
         		\specialrule{0em}{1pt}{1pt}
    			$F_{255}^1:\omega_1\omega_2\omega_3\omega_4\omega_5\omega_6\omega_7\omega_8$  &  0.1000  &0.0976 &0.0962& 0.0972&0.1026  \\
         		\specialrule{0em}{1pt}{1pt}
    			$F_{511}^1:\omega_1\omega_2\omega_3\omega_4\omega_5\omega_6\omega_7\omega_8\omega_9$  &  0.1000  &0.0976 &0.0962&  0.0973&  0.1029   \\
         		\specialrule{0em}{1pt}{1pt}
    			$F_{1023}^1:\omega_1\omega_2\omega_3\omega_4\omega_5\omega_6\omega_7\omega_8\omega_9\omega_{10}$  & 0.1000 & 0.0976  &0.0962 &0.0973 &  0.1029 \\
                   \specialrule{0em}{1pt}{1pt}
    			 \ \ \ \ \ \ \ \ \ \ \ \ \ \ \ \ \ \ $\vdots$ & \ \ \ \ $\vdots$  & \ \ \ \ $\vdots$&  \ \ \ \ $\vdots$&  \ \ \ \ $\vdots$& \ \ \ \ $\vdots$   \\
         		\specialrule{0em}{1pt}{1pt}
    			$F_{\infty}^1:\omega_1\omega_2\omega_3...\omega_{\infty}$  & 0.1000 & 0.0976  &0.0962 &0.0973 &  0.1029 \\
           		\toprule
                    &\multicolumn{5}{c}{$Conf$}  \\
                    \cmidrule(lr){2-6}
           			 \ \ \ \ \ \ \ \ \ \ \ \ \ \ \ \ \ $X$ & p=0.5 &p=0.6 & p=0.7 & p=0.8 &p=0.9\\
	            \midrule
           		\specialrule{0em}{1pt}{1pt}
    			$F_1^1: \omega_1$ & 0.2466 &0.2446  &0.2234 &	0.1786 & 0.1058  \\
        		\specialrule{0em}{1pt}{1pt}
    			$F_3^1:\omega_1\omega_2$ &0.1132 &0.1102 &0.1002& 0.0805& 0.0482  \\
        		\specialrule{0em}{1pt}{1pt}
    			$F_7^1:\omega_1\omega_2\omega_3$  & 0.0799  &0.0718 &0.0610 &  0.0464& 0.0266  \\
        		\specialrule{0em}{1pt}{1pt}
    			$F_{15}^1:\omega_1\omega_2\omega_3\omega_4$  & 0.0924 &0.0891  &0.0816&  0.0669& 0.0412  \\
        		\specialrule{0em}{1pt}{1pt}
    			$F_{31}^1:\omega_1\omega_2\omega_3\omega_4\omega_5$  & 0.1024  &0.1057 &0.1046&  0.0931 &0.0622 \\
         		\specialrule{0em}{1pt}{1pt}
    			$F_{63}^1:\omega_1\omega_2\omega_3\omega_4\omega_5\omega_6$  & 0.1086 &0.1181  &0.1248&  0.1193& 0.0858 \\
         		\specialrule{0em}{1pt}{1pt}
    			$F_{127}^1:\omega_1\omega_2\omega_3\omega_4\omega_5\omega_6\omega_7$  & 0.1122  &0.1267&0.1409& 0.1432&0.1101  \\
         		\specialrule{0em}{1pt}{1pt}
    			$F_{255}^1:\omega_1\omega_2\omega_3\omega_4\omega_5\omega_6\omega_7\omega_8$  & 0.1142  &0.1323 &0.1533&  0.1642& 0.1340  \\
         		\specialrule{0em}{1pt}{1pt}
    			$F_{511}^1:\omega_1\omega_2\omega_3\omega_4\omega_5\omega_6\omega_7\omega_8\omega_9$  & 0.1152  &0.1359 &0.1625&  0.1821 & 0.1570 \\
         		\specialrule{0em}{1pt}{1pt}
    			$F_{1023}^1:\omega_1\omega_2\omega_3\omega_4\omega_5\omega_6\omega_7\omega_8\omega_9\omega_{10}$  &0.1158 & 0.1381 &0.1693 & 0.1971 &0.1787   \\
                    \specialrule{0em}{1pt}{1pt}
    			\ \ \ \ \ \ \ \ \ \ \ \ \ \ \ \ \ \ $\vdots$ & \ \ \ \ $\vdots$  & \ \ \ \ $\vdots$&  \ \ \ \ $\vdots$&  \ \ \ \ $\vdots$& \ \ \ \ $\vdots$   \\
         		\specialrule{0em}{1pt}{1pt}
    			$F_{\infty}^1:\omega_1\omega_2\omega_3...\omega_{\infty}$  & 0.1163 & 0.1418  &0.1866 &0.2647 &0.4086 \\
    			\bottomrule
    			\label{tab7}
    		\end{tabular}
    	\end{center}
    \end{table}

It can be seen that the smaller $p$ is, the faster the weight decreases and reaches near 0 at a shallower depth, which leads to the conflict between $RPS_1$ and $RPS_2$ getting a convergence and stable state more quickly. When $p=0.1$, $Conf$ converges at a small depth of 4; when $p=0.2$, $ Conf$ converges at a depth of 7; whereas when $p=0.9$, $Conf$ does not converge until the depth is about 100. For the limit value $p=0$, it means that only the first element of the decision-maker's belief transfer propensity is considered, so no matter what $X$ is taken, there is a total conflict exists in the ordered focal sets $F_8^1$ and $F_7^1$, as well as $ F_5^1$ and $F_7^1$, non-conflict exists in $X$ and $F_7^1$. At this time, the conflict value between $RPS_1$ and $RPS_2$ is $1\times0.05\times1+1\times0.05\times 1=0.1$. On the other hand, the larger the $p$ is, the weight decreases relatively slowly, and the weights assigned to each depth are more even, which makes inconsistencies between lower-ranked elements still contribute to conflicts between RPSs. For the limit value $p=1$, the weight at any depth is 0, and $Conf$ is also 0, which is meaningless. Therefore, we do not present it in Table \ref{tab7}.

In addition, an attractive interpretation is that the parameter $p$ models the persistence of the decision-maker. Consider that the decision-maker always starts looking at the target element ranked first in qualitative propensity. At each depth, they have a probability $p$ of continuing to the next ranked element, and the probability of deciding to stop is $1-p$. Once decision-makers lose patience at depth $d$, they calculate the degree of inconsistency between the two RPSs at that depth and use it as a measure of conflict between RPSs. Regardless, the parameter $p$ provides the decision-maker with an option that allows the decision-maker to adjust the weight of the top-ranked elements according to preferences or practical application requirements, proving the flexibility of the proposed conflict measure method.

\section{Conclusion}
\label{5}
This paper discusses a novel order-sensitive conflict measure for RPST from the perspective of DST. First, we analyze various conflict scenarios in RPS from the viewpoints of RFS and DST, and point out the limitations of existing approaches for measuring conflicts in RPS. From the DST perspective, an essential characteristic of ordered focal sets is that higher-ranked elements occupy a more critical position. Based on the observation of simple permutations, we first define the notion of conflict between permutations and then propose an order-sensitive conflict measure for PMFs in RPS. It characterizes conflict in RPSs as a graded, order-sensitive notion, moving beyond the traditional criterion that determines conflict merely by the existence of an intersection. Several numerical examples are used to demonstrate the properties and behavior of the proposed measure. The proposed conflict measure can not only effectively measure the conflicts between PMFs but also has the natural top-weightedness property, as well as the unique potential of arbitrary truncation, adjustable parameters, and measure conflicts between RPSs with infinite depth. The proposed conflict measure provides a useful tool for subsequent fusion of uncertain information with order information and for managing conflicts throughout the fusion process, which also constitutes a key direction of our future research.

\section*{CRediT authorship contribution statement}
\textbf{Ruolan Cheng}: Conceptualization, Methodology, Formal analysis, Software, Visualization, Validation, Investigation, Writing - original draft, Writing - review \& editing. \textbf{Yong Deng}: Resources, Supervision, Funding acquisition, Writing - review \& editing.

\section*{Declaration of Competing Interest}
The authors declare that they have no known competing financial interests or personal relationships that could have appeared to influence the work reported in this paper.

\section*{Acknowledgment}
This work is supported by the National Natural Science Foundation of China (Grant No. 62373078) and the China Scholarship Council (202306070060).

\bibliography{References}
\bibliographystyle{elsarticle-num}

\end{document}